\documentclass[10pt,twocolumn,letterpaper]{article}

\usepackage[pagenumbers]{cvpr} 

\PassOptionsToPackage{hypertexnames=false}{hyperref}
\usepackage{multirow}
\usepackage{tabularx}
\usepackage{array}
\usepackage{makecell}
\usepackage{adjustbox}
\usepackage{capt-of}
\usepackage{placeins}
\usepackage{needspace}
\usepackage{ragged2e}
\usepackage{soul}
\usepackage{CJKutf8}
\newcommand{\cjkchar}[1]{\begin{CJK}{UTF8}{gbsn}#1\end{CJK}}
\newcommand{\best}[1]{\textbf{#1}}
\newcommand{\second}[1]{\underline{#1}}
\newcolumntype{Y}{>{\centering\arraybackslash}X}   
\newenvironment{fitwidthtabular}[1]
  {\begin{adjustbox}{max width=\linewidth}\begin{tabular}{#1}}
  {\end{tabular}\end{adjustbox}}

\usepackage{colortbl}
\usepackage{tikz}
\definecolor{tableblue}{HTML}{527DA5}
\definecolor{tableterra}{HTML}{C47761}
\definecolor{tableink}{HTML}{48515B}
\definecolor{tablerule}{HTML}{AAB5BF}
\definecolor{oursrow}{HTML}{E9F0F6}
\definecolor{headrow}{HTML}{F0F3F6}
\definecolor{keycol}{HTML}{F5F8FB}
\definecolor{mutedtext}{HTML}{707B85}
\newcommand{\tablestyle}{%
  \renewcommand{\arraystretch}{1.08}%
  \setlength{\aboverulesep}{1.8pt}%
  \setlength{\belowrulesep}{1.8pt}%
  \setlength{\heavyrulewidth}{0.65pt}%
  \setlength{\lightrulewidth}{0.35pt}%
  \arrayrulecolor{tablerule}%
}
\AtBeginEnvironment{table}{\tablestyle}
\AtBeginEnvironment{table*}{\tablestyle}
\newcommand{\tableheading}[1]{\textcolor{tableink}{\textbf{#1}}}

\newcommand{\ours}{\rowcolor{oursrow}}       
\newcommand{\headband}{\rowcolor{headrow}}   
\newcommand{\na}{\textcolor{mutedtext}{--}}  
\newcommand{\mods}[1]{\textcolor{mutedtext}{\scriptsize #1}} 
\newcommand{\ablationbar}[1]{%
  \tikz[x=1pt,y=1pt,baseline=0pt]{%
    \fill[black!10,rounded corners=0.7pt] (0,0) rectangle (17,3.4);%
    \pgfmathsetmacro{\barlength}{17*abs(#1)/6.02}%
    \ifdim\barlength pt>0pt
      \fill[tableblue,rounded corners=0.7pt] (0,0) rectangle (\barlength,3.4);%
    \fi
  }%
}
\newcommand{\dbar}[1]{%
  \ablationbar{#1}\hspace{2pt}%
  \makebox[18pt][r]{\textcolor{tableterra}{\scriptsize $#1$}}%
}
\newcommand{\dbarfull}{%
  \ablationbar{0}\hspace{2pt}%
  \makebox[18pt][r]{\textcolor{tableblue}{\scriptsize 0.00}}%
}
\newcommand{\group}[2]{\multicolumn{#1}{@{}l}{\textcolor{tableblue}{\textbf{#2}}}} 
\newcommand{\APb}{AP$_{\text{base}}$}
\newcommand{\APf}{AP$_{\text{few}}$}
\newcommand{\APz}{AP$_{\text{zero}}$}
\newcommand{\Hbf}{$H_{\text{b,f}}$}
\newcommand{\Hbz}{$H_{\text{b,z}}$}

\makeatletter\@clubpenalty=10000\makeatother
\AtBeginDocument{\parfillskip=0pt plus 0.8\columnwidth}
\DeclareCaptionJustification{justified}{\parfillskip=0pt plus 1fil}

\newcommand{\paperdisplaysep}{%
  \setlength{\abovedisplayskip}{5.5pt plus 1pt minus 1pt}%
  \setlength{\belowdisplayskip}{5.5pt plus 1pt minus 1pt}%
  \setlength{\abovedisplayshortskip}{2pt plus 1pt minus 1pt}%
  \setlength{\belowdisplayshortskip}{3pt plus 1pt minus 1pt}%
}
\AtBeginDocument{%
  \apptocmd{\normalsize}{\paperdisplaysep}{}%
    {\PackageError{Struct2Det}{Could not set equation spacing}{}}%
  \paperdisplaysep
}

\usepackage[ruled,lined,linesnumbered]{algorithm2e}
\SetAlCapFnt{\bfseries}
\SetAlCapNameFnt{\normalfont}
\SetAlCapSkip{3pt}
\SetAlgoCaptionSeparator{\quad}
\SetKwInput{KwInput}{Input}
\SetKwInput{KwOutput}{Output}
\SetKw{Break}{break}
\SetAlgoNlRelativeSize{-1}
\SetNlSty{textnormal}{}{}
\SetInd{0.35em}{0.55em}
\SetAlFnt{\footnotesize}

\newcommand{\loss}[1]{\mathcal L_{\mathrm{#1}}}
\newcommand{\op}[1]{\operatorname{#1}}

\usepackage{microtype}
\makeatletter
\patchcmd{\abstract}{\vspace*{12pt}\noindent}{\vspace*{3pt}\noindent}{}{}
\makeatother

\usepackage{bbding} 

\definecolor{cvprblue}{rgb}{0.21,0.49,0.74}
\usepackage[pagebackref,breaklinks,colorlinks,allcolors=cvprblue]{hyperref}

\def\confName{CVPR}
\def\confYear{2027}

\title{Struct2Det: Grounding Category Structure in Visual Evidence \\
for Few-Shot Object Detection}

\author{
    Kaixuan Chang$^1$ \quad
    Huabing Zhou$^{1~ \text{\Envelope}}$
    \quad
    Yanduo Zhang$^1$ \quad
    Tao Lu$^1$ \\
    Liwei Wang$^1$ \quad
    Zhongyuan Wang$^2$ \quad
    Jiayi Ma$^2$ \\
    {\small $^1$Hubei Key Laboratory of Intelligent Robot, Wuhan Institute of Technology} \quad
    {\small $^2$Wuhan University} \\
}

\makeatletter
\ifboolexpr{togl {cvprfinal} and togl {cvprpagenumbers}}{%
    \patchcmd{\@maketitle}{\vskip .375in}{\vskip .25in}{}%
        {\PackageError{Struct2Det}{Could not adjust arXiv title spacing}{}}%
    \patchcmd{\@maketitle}{\vspace*{24pt}}{\vspace*{8pt}}{}%
        {\PackageError{Struct2Det}{Could not adjust arXiv title spacing}{}}%
    \patchcmd{\@maketitle}{\vspace*{12pt}}{\vspace*{4pt}}{}%
        {\PackageError{Struct2Det}{Could not adjust arXiv title spacing}{}}%
    \AtBeginDocument{\raggedbottom}%
}{}
\makeatother

\begin{document}
\maketitle
\addtocontents{toc}{\protect\setcounter{tocdepth}{-2}}
\begin{abstract}
Few-shot detectors struggle to distinguish categories that look alike but differ in local components, because a few examples rarely reveal which differences determine category identity. Category descriptions specify components and relations but do not establish their presence in an image. We propose \emph{Struct2Det}, which uses category structure to guide detection and verify candidates in three stages. \emph{Relational Structure Encoding} maps category graphs to component embeddings and a global condition. \emph{Component-Grounded Detection} uses component queries to retrieve multi-scale image evidence that, with the global condition, modulates category text embeddings for initial detection. \emph{Verification-Guided Refinement} matches candidate regions with components expected by their top-scoring categories; when class scores are close and matches are reliable, matched pairs correct region features before the shared detector predicts again. Category structure informs both classification inputs: text embeddings integrate image-wide component evidence, while region features incorporate candidate-specific matches. We also construct \emph{InduSym}, with 88,912 symbol instances across 275 categories from 674 engineering drawings and a structural graph per category. Struct2Det improves few-shot-category AP over the strongest baseline by 6.02 points on InduSym and 9.68 on Digitize-PID, with the highest base-category AP on both. InduSym ablations show few-shot gains from correct category structure, image-level retrieval, and candidate verification. Further tests show gains in few-shot natural-object detection and zero-shot Chinese character recognition.
\end{abstract}
    
\section{Introduction}
\label{sec:intro}

\begin{figure}[!t]
  \centering
  \includegraphics[width=\linewidth]{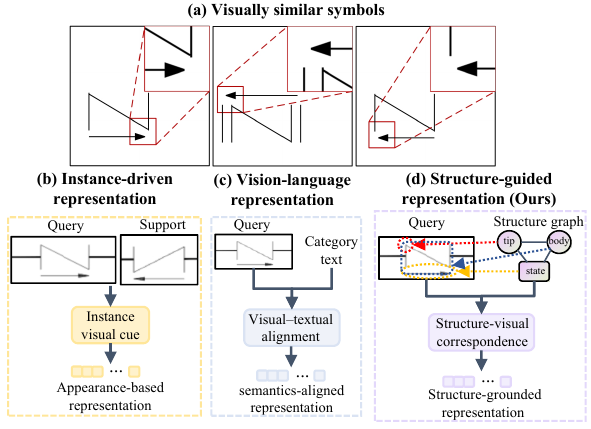}
  \caption{\textbf{From appearance to structure.} (a) Similar symbols with local differences highlighted in red; (b) instance-driven appearance representation; (c) vision-language semantic representation; and (d) our structure-guided representation with component matching and verification ($\checkmark$).}
  \label{fig:performance}
\end{figure}

Few-shot object detection aims to detect objects of novel categories given only a few annotated instances~\cite{kang2019fewshot,wang2020frustratingly}. A particular challenge arises when categories look alike overall but differ in local components. In industrial symbols, for example, a short line, a port, or an additional enclosure can change a symbol's meaning while leaving most of its appearance unchanged (Figure~\label{firstref:fig:performance}\ref{fig:performance}a)~\cite{paliwal2021digitizepid,jamieson2024review}. A few instances rarely reveal which of these local differences determine category identity and which are incidental appearance variation.

Category descriptions often supply the knowledge that a few instances fail to reveal. Engineering legends~\cite{jamieson2024review}, ideographic description sequences~\cite{zeng2023star}, and category definitions specify components and relations independently of instance annotations, identifying which local differences define a category. For industrial symbols, the legend defines how standardized graphical primitives~\cite{iso10628part2,isa51} are arranged and connected; this structure remains stable across drawings despite appearance variation.

Figure~\ref{fig:performance} contrasts two common sources of category information. Few-shot methods learn appearance cues from support examples~\cite{yu2026acloser,kang2019fewshot,wang2020frustratingly,yan2019meta}, whose coverage is limited by the available instances. Vision-language detectors align image regions with category text embeddings~\cite{wu2023cora,liu2024grounding,cheng2024yoloworld}, but this region-level compatibility leaves component correspondence implicit. Neither representation provides an explicit check of expected components within a predicted candidate.

Knowing which components a category should contain does not establish their presence in an image. Evidence of these components can first be gathered across the image to guide detection. Yet a candidate's class score reflects overall region--category compatibility and can remain high even when a distinguishing component is absent. Candidates must therefore also be checked for expected components within their regions.

To this end, we introduce Struct2Det, which uses category structure to guide detection and verify candidates (Figure~\ref{fig:performance}d). Its three stages operate at the category, image, and candidate levels. Each description becomes a category structural graph (CSG) with component nodes carrying geometric attributes and edges encoding typed relations. At the category level, Relational Structure Encoding maps the CSG to component embeddings and a global condition summarizing its structure, which together specify \emph{what} components are expected. At the image level, Component-Grounded Detection uses the component embeddings to query multi-scale image features for evidence of \emph{where} these components appear. This evidence and the global condition jointly modulate the category text embeddings used by the detector to predict initial boxes and class scores.

At the candidate level, Verification-Guided Refinement applies node--region matching to check \emph{whether} each candidate region contains components expected by its top-scoring categories. When the top two class scores are close and the structural matches are reliable, a correction derived from the matched pairs updates the candidate's region features, and the same detector predicts again. Category structure is thus grounded in two ways: image-wide component evidence modulates the category text embeddings, while candidate-specific matches correct the region features. These representations form the two inputs to the classification head.

Industrial symbols provide a natural test bed for this problem because categories differ in local structure and their frequencies follow a long-tailed distribution~\cite{ma2026artificial,jamieson2024review,byun2025pid}. We construct InduSym, a benchmark of 88,912 symbol instances across 275 categories (204 base, 45 few-shot, and 26 zero-shot) from 674 engineering drawings, with a structural graph for every category. A script parses each SVG legend symbol into a graph, which domain experts verify and correct.

The main contributions are:
\begin{itemize}

\item We propose Struct2Det, which combines image-wide component retrieval for category conditioning with candidate-level verification, in which reliable node--region matches correct the region features of ambiguous candidates.

\item We construct InduSym, an industrial-symbol detection benchmark that pairs instance annotations with a structural graph for every category. The benchmark and our code will be released.

\item On InduSym and Digitize-PID, Struct2Det raises few-shot AP by 6.02 and 9.68 points over the strongest baseline while also achieving the highest base AP. Ablations assess correct component relations, image-level retrieval, and candidate matching with local feedback. Further evaluations show gains in few-shot natural-object detection and zero-shot Chinese character recognition.
\end{itemize}

\section{Related Work}

\textbf{Category Information in Few-Shot and Open-Vocabulary Detection.}
Few-shot object detection adapts a detector to novel categories from a small support set. Representative methods model category cues through feature reweighting and meta-learning~\cite{kang2019fewshot,yan2019meta}, fine-tuning~\cite{wang2020frustratingly,yu2026acloser}, contrastive proposal encoding~\cite{sun2021fsce}, decoupled classification and localization~\cite{qiao2021defrcn}, and support-query feature interaction~\cite{han2022crosstransformer,han2026multimodal}. Their category cues come from support features, so coverage is limited by the available instances and the defining components remain implicit. Language offers a second source of category information. Open-vocabulary detectors align visual regions with textual concepts~\cite{radford2021clip,zareian2021open,minderer2022owlvit}, and refine this alignment through region-level pretraining~\cite{zhong2022regionclip,kim2023regionaware,wu2023cora}, distillation~\cite{gu2022open,ma2022open,fu2025hdovd}, concept expansion~\cite{zhou2022detic,yao2022detclip,feng2022promptdet}, and grounded pretraining~\cite{zhang2022glipv2,liu2024grounding}. YOLO-World~\cite{cheng2024yoloworld} and YOLOE~\cite{wang2025yoloe} make this alignment efficient. LLMDet~\cite{fu2025llmdet} uses generated image-level and region-level captions as supervision. Such alignment does not explicitly test component correspondence within a candidate. Building on YOLO-World, we use category graphs for component queries and candidate-level matching.

\begin{figure*}[!t]
\centering
\includegraphics[width=0.94\textwidth]{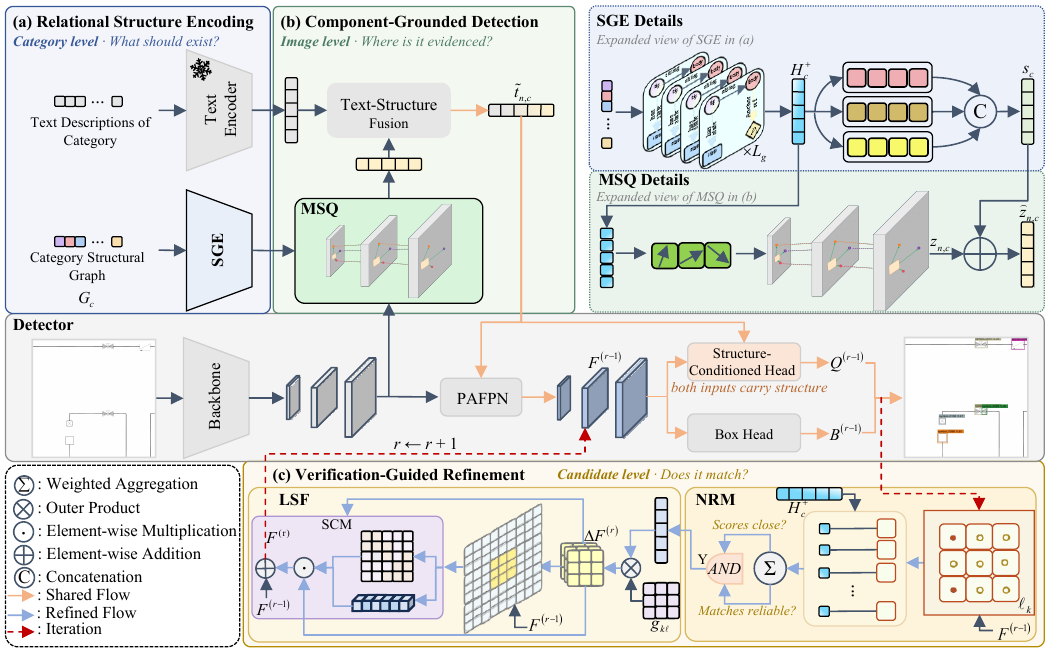}
\caption{\textbf{Overall architecture of Struct2Det.} (a) Relational Structure Encoding specifies \emph{what} components to expect (category level). (b) Component-Grounded Detection retrieves evidence of \emph{where} they appear to modulate text embeddings (image level). (c) Verification-Guided Refinement checks \emph{whether} candidates contain expected components and corrects their region features (candidate level). Both classification inputs, category embeddings and region features, carry structural evidence. The detector is shared across rounds.}
\label{fig:framework}
\end{figure*}

\smallskip
\noindent\textbf{Parts and Structure in Visual Recognition.}
Structured representations have been used across visual recognition. Part-based models decompose an object into parts and model their spatial configuration~\cite{felzenszwalb2010object}. Scene graph methods represent entities and relations~\cite{li2022sgtr,wang2025entitypredicate}. Document understanding methods model text, visual content, and two-dimensional layout~\cite{huang2022layoutlmv3,appalaraju2021docformer,tang2023udop}. Character recognition methods exploit components, spatial composition, and structural constraints~\cite{chen2021zeroshotchar,chen2022acpm,li2025oraclefusion}. Industrial symbol recognition similarly relies on the composition of standardized parts and their geometric arrangements to distinguish visually similar symbols~\cite{kim2022endtoend,gupta2024ssdpid}. Recent graph-structured models recover connectivity between drawing elements for diagram digitization~\cite{hu2026pidcon,sturmer2025pid2graph}, a different structural level from the composition within an individual symbol. These methods typically learn parts from instances or recover connectivity from drawings, whereas our parts come from category descriptions and our graph grounds detection in image evidence.

\smallskip
\noindent\textbf{Iterative Refinement and Verification.}
A third line of work concerns how predictions are refined and verified. Query-based detectors use learned queries or proposals for multi-stage refinement~\cite{carion2020end,zhu2021deformable,sun2021sparse}. Template matching and regression~\cite{jo2025fewshotpattern} retains exemplar spatial layout to match and refine candidates. Neither verifies whether a candidate actually contains the structure that its top-scoring category requires. Our structural prior instead comes from category descriptions, and local component matches compare competing category hypotheses before feature correction. We adopt deformable sampling~\cite{zhu2021deformable} for image-level retrieval; candidate-level matching supplies the separate local check.

\section{Method}\label{sec:method}

\subsection{Overall Framework}

Given an image $I_n$ and a category structural graph (CSG) $G_c$ for each category $c$, Struct2Det detects objects in three stages, each addressing one question (Figure~\label{firstref:fig:framework}\ref{fig:framework}). Relational Structure Encoding maps $G_c$ to component embeddings $\mathbf H_c^+$ and a global condition $\mathbf s_c$, specifying \emph{what} components category $c$ should contain. Component-Grounded Detection uses $\mathbf H_c^+$ to query multi-scale image features for evidence of \emph{where} these components appear. This evidence and $\mathbf s_c$ modulate the category text embeddings into $\widetilde{\mathbf t}_{n,c}$, with which the detector produces initial features $\mathbf F_n^{(0)}$, boxes $\mathbf B^{(0)}$, and scores $\mathbf Q^{(0)}$. Verification-Guided Refinement checks \emph{whether} each candidate region contains components expected by its top-scoring categories. In round $r$, reliable matches correct ambiguous candidates' features to yield $\mathbf F_n^{(r)}$, from which shared heads predict $\mathbf B^{(r)}$ and $\mathbf Q^{(r)}$. At initial detection and each refinement round, the structure-conditioned classification head compares region features from $\mathbf F_n^{(r)}$ with category embeddings $\widetilde{\mathbf t}_{n,c}$. Both inputs thus carry category structure: text embeddings integrate image-wide component evidence, while updated region features incorporate candidate-specific matches.

\subsection{Category Structural Graph}

We represent each category $c$ as an attributed graph $G_c=(\mathcal V_c,\mathcal E_c,\mathbf X_c,\mathbf R_c)$, where $\mathcal V_c$ and $\mathcal E_c$ are the node and edge sets, and $\mathbf X_c$ and $\mathbf R_c$ contain their attributes. In every domain, the same parser constructs $G_c$ from a structured SVG. Nested groups specify the hierarchy, primitives describe component geometry, and anchors mark connection locations; coordinates are normalized using the root \texttt{viewBox}.

InduSym provides an SVG for every category. For the other datasets, Qwen3-VL-Plus~\cite{qwen3vl2025} generates the SVG from the category's legend description (Digitize-PID), ideographic description sequence (Chinese characters), or public definition (natural objects). A representative prompt, parsing details, and graph examples are given in the Supplementary Material.

\subsection{Relational Structure Encoding}

The Structure Graph Encoder (SGE) maps $G_c$ to component embeddings $\mathbf H_c^+$ and a global condition $\mathbf s_c$. The attributes of node $i$ are first embedded into $\mathbb R^d$, and $L_{\mathrm g}$ relation-aware message-passing layers~\cite{gilmer2017neural} then update this embedding, giving $\mathbf h_i^{(\tau)}$ after layer $\tau$. Each message combines a neighboring node's embedding with the relation embedding of the connecting edge, and a residual update integrates the aggregated messages with the node's previous state $\mathbf h_i^{(\tau-1)}$. Relations of different types, such as containment and connection, thus contribute differently to each component's representation.

$\mathbf H_c^+$ stacks the final embeddings $\mathbf h_{c,i}^{+}=\mathbf h_i^{(L_{\mathrm g})}$ of the nodes that represent visual components, such as primitives, parts, or radicals; only roots and groups are excluded. For categories with different component counts, three readouts produce a fixed-dimensional condition:
\begin{equation}
\mathbf s_c=\phi_{\mathrm o}\!\left([\mathbf g_c;\mathbf p_c;\mathbf r_c]\right),
\end{equation}
where $[\cdot;\cdot]$ concatenates and $\phi_{\mathrm o}$ is a two-layer MLP with output in $\mathbb R^d$. The readouts combine salience-weighted pooling ($\mathbf g_c$), attention to $P$ learned probes~\cite{lee2019settransformer} ($\mathbf p_c$), and element-wise maxima ($\mathbf r_c$). Thus $\mathbf s_c$ summarizes category structure, while $\mathbf H_c^+$ retains individual components for retrieval and verification. Neither indicates where components appear in an image.

\subsection{Component-Grounded Detection}

To find image evidence for the expected components, the Multi-Scale Structural Query (MSQ) uses each component embedding $\mathbf h_{c,i}^{+}$, a row of $\mathbf H_c^{+}$, as a query over multi-scale features. Each query predicts a reference point in normalized image coordinates, $K$ sampling offsets per attention head and feature level, and normalized reading weights. Multi-scale deformable sampling~\cite{zhu2021deformable} then reads the detector~\cite{cheng2024yoloworld} backbone features $\mathbf Y_{n,\ell}$ of image $I_n$, $\ell\in\{3,4,5\}$, at these locations. Components are queried separately; their responses are then smoothed over the graph adjacency and mean-pooled into the image evidence $\mathbf z_{n,c}\in\mathbb R^d$. Because sampling locations are predicted from the component embedding alone, this evidence is image-level and shared by all instances of category $c$; the detector localizes instances, and Sec.~\ref{sec:vgr} checks components inside candidates.

Within MSQ, this evidence is combined with the global condition, and the result modulates the category text embedding:
\begin{equation}
\begin{aligned}
\hat{\mathbf z}_{n,c}&=\operatorname{LN}\!\left(\mathbf s_c+\alpha\,\operatorname{proj}(\mathbf z_{n,c})\right),\\
\widetilde{\mathbf t}_{n,c}&=\operatorname{LN}\!\left(\mathbf t_{n,c}+\lambda\,\psi\!\left(\phi_{\mathrm{fuse}}(\mathbf t_{n,c},\hat{\mathbf z}_{n,c})\right)\right).
\end{aligned}
\end{equation}
Here $\mathbf t_{n,c}$ comes from the detector's CLIP text encoder, $\operatorname{proj}$ is a linear projection, and $\operatorname{LN}$ denotes LayerNorm. The nonnegative learnable weight $\alpha$ starts at zero, so $\hat{\mathbf z}_{n,c}$ initially retains only the global condition. The element-wise gated residual fusion $\phi_{\mathrm{fuse}}$ and the two-layer adapter $\psi$ combine text and structural evidence, with $\lambda$ scaling the adapter output. Their definitions are in the Supplementary Material.

The modulated embeddings $\widetilde{\mathbf t}_{n,c}$ condition the PAFPN and replace the class embeddings that the classification head scores against. The detector produces initial features $\mathbf F_n^{(0)}$, boxes $\mathbf B^{(0)}$, and scores $\mathbf Q^{(0)}$; the image index of $\mathbf B$ and $\mathbf Q$ is omitted. For initial detection, structure enters through these embeddings; the head architecture is unchanged.

\subsection{Verification-Guided Refinement}\label{sec:vgr}

Image-wide component evidence does not directly establish component presence inside each candidate. A class score measures how well a region fits a category as a whole; it cannot tell whether the one component that separates two look-alike categories is present. We therefore compare each candidate's top-scoring categories, treated as hypotheses, through local component matches. Better-matched hypotheses guide a region-feature correction, after which the shared detector predicts again. At inference, only ambiguous candidates with reliable visual--structural matches are corrected, for up to $R_{\max}$ rounds. Below, we fix image $I_n$ and omit the index $n$ from \mbox{candidate quantities}.

\smallskip
\noindent\textbf{Node--region matching (NRM).} NRM establishes online visual--structural correspondence for each candidate. At round $r$, it samples features $\mathbf F_{n,\ell}^{(r-1)}$ on a grid inside the previous box $\mathbf B_k^{(r-1)}$, using the level $\ell$ that produced candidate $k$. The resulting visual tokens $\mathbf f_{k,q}^{(r)}$ and component embeddings $\mathbf h_{c,i}^{+}$ of hypotheses $c\in\mathcal C_k^{(r)}$ are projected to a common embedding space. Their pairwise match score is $s_{k,c,q,i}^{(r)}=\tfrac12[1+\tanh(\cos(\mathbf f_{k,q}^{(r)},\mathbf h_{c,i}^{+}))]$.

For each of the top five category hypotheses $c\in\mathcal C_k^{(r)}$, NRM first selects the four highest-scoring pairs $(q,i)$. Pairs whose grid cell and component are not each other's best match are down-weighted, giving the adjusted score $\bar s_{k,c,q,i}^{(r)}$. Thresholding this score and filtering spatial and graph-based redundancy leave the retained correspondence set $\mathcal I_{k,c}^{(r)}$. Matching seeks supporting component evidence; it does not check every component or explicitly verify all spatial relations. Relations inform both node embeddings and \mbox{redundancy filtering}.

\smallskip
\noindent\textbf{Structural correction.} Inspired by expert mixtures~\cite{shazeer2017moe}, a shared low-rank module $E_\theta$ uses each retained pair's visual token, component embedding, and adjusted score. We first average its corrections within each hypothesis, then aggregate across hypotheses:
\begin{equation}
\begin{aligned}
\Delta\mathbf u_{k,c}^{(r)}
&=\frac{\sum_{(q,i)\in\mathcal I_{k,c}^{(r)}}\bar s_{k,c,q,i}^{(r)}E_\theta(\mathbf f_{k,q}^{(r)},\mathbf h_{c,i}^{+},\bar s_{k,c,q,i}^{(r)})}{\sum_{(q,i)\in\mathcal I_{k,c}^{(r)}}\bar s_{k,c,q,i}^{(r)}+\epsilon},\\
\Delta\mathbf u_k^{(r)}&=\sum_{c\in\mathcal C_k^{(r)}}a_{k,c}^{(r)}\Delta\mathbf u_{k,c}^{(r)}.
\end{aligned}
\end{equation}
Here $\epsilon>0$ stabilizes normalization. The weights $a_{k,c}^{(r)}$ are a softmax over the log detector score plus a weighted strongest-match score. Stronger matches thus increase a hypothesis's weight at a given detector score. Categories without retained pairs receive zero weight; if none remain, correction and reliability are zero.

\smallskip
\noindent\textbf{Reliability gate.} A correction is applied only if it is both needed and supported. Need is measured by ambiguity: $m_k^{(r)}$ is the gap between candidate $k$'s highest and second-highest class scores in $\mathbf Q^{(r-1)}$. Support is measured by reliability, the strongest retained match of each category hypothesis averaged with the category weights:
\begin{equation}
\rho_k^{(r)}=\sum_{c\in\mathcal C_k^{(r)}}a_{k,c}^{(r)}\max_{(q,i)\in\mathcal I_{k,c}^{(r)}}\bar s_{k,c,q,i}^{(r)},
\end{equation}
where the maximum over an empty set is zero. At inference, the correction is accepted when
\begin{equation}
\Gamma_k^{(r)}=\mathbf 1\!\left[m_k^{(r)}<\delta_m^{(r)},\quad\rho_k^{(r)}\ge\delta_\rho^{(r)}\right],
\end{equation}
with round-specific thresholds $\delta_m^{(r)}$ and $\delta_\rho^{(r)}$. All sizes, thresholds, and training settings are given in the Supplementary Material.

\smallskip
\noindent\textbf{Local structural feedback (LSF).} LSF writes each accepted correction back to the feature map that produced its candidate. At location $p$ on level $\ell$, overlapping corrections are combined by weighted averaging:
\begin{equation}
\Delta\mathbf F_{n,\ell}^{(r)}(p)=\mathbf W_\ell
\frac{\sum_{k\in\mathcal K_\ell^{(r)}}w_{k\ell}^{(r)}(p)\,\Delta\mathbf u_k^{(r)}}{\sum_{k\in\mathcal K_\ell^{(r)}}w_{k\ell}^{(r)}(p)+\epsilon},
\end{equation}
where $\mathcal K_\ell^{(r)}$ contains candidates that pass the reliability gate ($\Gamma_k^{(r)}=1$) and originate from pyramid level $\ell$. The learned $1\times1$ projection $\mathbf W_\ell$ maps corrections to that level's channels, and $w_{k\ell}^{(r)}(p)=\rho_k^{(r)}\mathbf{g}_{k\ell}(p)$ combines reliability with local support. Here $\mathbf{g}_{k\ell}(p)$ is the scalar Gaussian weight at location $p$, supported on a fixed $3\times3$ neighborhood around the candidate's previous-round box center on level $\ell$.

NRM determines the correction; LSF controls its local support and write strength, largely preserving features elsewhere. Within LSF, Spatial--Channel Modulation (SCM) helps suppress unintended updates to neighboring objects and unrelated channels through two learned convolutional networks. A spatial network computes one sigmoid weight per location from current features, residual, and locally scattered reliability $\rho_k^{(r)}$ and ambiguity $1-m_k^{(r)}$ maps. A channel network computes one sigmoid weight per channel from globally averaged features and residual. Their broadcast product forms $\mathbf M_{n,\ell}^{(r)}$. Unlike the dimension-changing $\mathbf W_\ell$, the $1\times1$ projection $\mathbf P_\ell$ mixes residual channels; its zero initialization preserves the initial detector features:
\begin{equation}
\mathbf F_{n,\ell}^{(r)}=\mathbf F_{n,\ell}^{(r-1)}+\operatorname{clip}_{\beta}\!\left(\eta_\ell\mathbf M_{n,\ell}^{(r)}\odot\mathbf P_\ell(\Delta\mathbf F_{n,\ell}^{(r)})\right),
\end{equation}
where $\eta_\ell$ is a learned per-level scale, $\odot$ denotes element-wise multiplication, and $\operatorname{clip}_{\beta}$ rescales an update if its root-mean-square (RMS) value exceeds $\beta$ times that of the pre-update features.

After LSF updates the features, the shared detection heads re-predict boxes $\mathbf B^{(r)}$ and class scores $\mathbf Q^{(r)}$, replacing only accepted candidates' predictions. Matching then repeats with the updated category hypotheses and boxes. Refinement stops when no candidate passes the gate or after $R_{\max}$ rounds, returning the last predictions.

\subsection{Training Objective}\label{sec:objective}

The main objective combines detection supervision with three structural terms:
\begin{equation}
\mathcal L=\loss{det}+\loss{cond}+\frac{1}{\max(1,R)}\sum_{r=1}^{R}\bigl(\loss{loc}^{(r)}+\lambda_{\mathrm{fb}}\loss{fb}^{(r)}\bigr),
\end{equation}
where $R\le R_{\max}$ counts executed matching rounds, and the weights of $\loss{cond}$ and $\loss{loc}$ are included in these terms. Each term constrains a different quantity.

\noindent\textbf{Detection.} $\loss{det}$ combines classification and box-regression losses for the initial prediction ($r=0$) and each refinement round that applies a correction, using scheduled weights for the refinement rounds.

\noindent\textbf{Conditioning consistency.} $\loss{cond}$ anchors the global condition $\mathbf s_c$ to a cached per-category reference $\bar{\mathbf s}_c$ and keeps the modulated embedding $\widetilde{\mathbf t}_{n,c}$ close to $\mathbf t_{n,c}$, limiting drift in category conditioning.

\noindent\textbf{Local correspondence.} $\loss{loc}^{(r)}$ acts on the adjusted match scores $\bar s_{k,c,q,i}^{(r)}$. For candidates whose category hypotheses include the ground-truth category, it raises the strongest match to that category and enforces a margin over the strongest match to any other hypothesis.

\noindent\textbf{Feedback magnitude.} $\loss{fb}^{(r)}$ penalizes the RMS ratio of the feature update to the pre-update features $\mathbf F_{n,\ell}^{(r-1)}$ where it exceeds $\beta$, measured before clipping, complementing SCM's spatial and channel gating.

\noindent\textbf{Few-shot adaptation.} Both phases enable refinement and freeze only the text encoder. Phase I uses base categories; Phase II adds few-shot support instances. GT-class injection, which adds a missing ground-truth class to a candidate's hypotheses, is training-only and decays to zero. CSG construction does not use support instances.

Auxiliary regularizers and backbone-specific training terms supplement this objective. Implementation details, weights, and the two-phase training procedure appear in the Supplementary Material.
\FloatBarrier

\section{Experiments}

\setlength{\intextsep}{4pt plus 2pt minus 2pt}

\begin{table*}[t]
\centering
\caption{\textbf{Industrial-symbol detection on InduSym and Digitize-PID.} Metrics follow Sec.~\ref{sec:BAS}; SSDPID's overall AP excludes zero-shot categories. Best in \textbf{bold}, second best \underline{underlined}; \na{}: unavailable; $\Delta$: Struct2Det minus the best baseline.}\label{tab:main-results}
\small
\setlength{\tabcolsep}{4.1pt}
\renewcommand{\arraystretch}{0.96}
\setlength{\aboverulesep}{1pt}
\setlength{\belowrulesep}{1pt}
\colorlet{fewcol}{tableterra!9}%
\newcommand{\dpos}[1]{\textcolor{tableblue}{\scriptsize +#1}}%
\newcommand{\dneg}[1]{\textcolor{tableterra}{\scriptsize $-$#1}}%
\begin{tabularx}{\textwidth}{l *{6}{Y} c Y >{\columncolor{fewcol}}Y *{3}{Y}}
\toprule
\multicolumn{1}{@{}l}{} & \multicolumn{6}{c}{\tableheading{Detection metrics}} & & \multicolumn{5}{c}{\tableheading{Category-wise}} \\
\cmidrule(lr){2-7}\cmidrule(l){9-13}
\headband \tableheading{Method} & AP & AP$_{50}$ & AP$_{75}$ & AP$_{s}$ & AP$_{m}$ & AP$_{l}$ & & \APb & \textbf{\APf} & \APz & $H_{\text{b,f}}$ & $H_{\text{b,z}}$ \\
\midrule
\group{13}{InduSym} \\
RegionCLIP & 23.05 & 32.67 & 26.25 & 6.87 & 18.28 & 30.53 &  & 27.20 & 20.15 & 0.00 & 23.15 & 0.01 \\
CORA & 3.69 & 9.20 & 2.36 & 0.18 & 1.53 & 8.88 &  & 4.50 & 2.65 & 0.00 & 3.34 & 0.00 \\
YOLO-World & 19.38 & 25.70 & 22.10 & 8.80 & 15.50 & 30.00 &  & 19.07 & \second{31.70} & 0.11 & 23.81 & 0.22 \\
Grounding DINO & 11.33 & 16.49 & 12.22 & 8.36 & 9.77 & 21.29 &  & 15.03 & 3.36 & 0.08 & 5.49 & 0.15 \\
SSDPID & 11.53 & 15.86 & 13.40 & 7.92 & 11.45 & 11.43 &  & 10.76 & 14.55 & \na & 12.37 & \na \\
YOLOE & 3.98 & 6.96 & 4.08 & 1.48 & 3.11 & 6.11 &  & 5.55 & 0.13 & 0.00 & 0.26 & 0.00 \\
HD-OVD & \second{26.92} & \second{36.90} & \second{31.00} & 12.10 & \second{24.20} & \second{32.00} &  & \second{32.57} & 20.35 & 0.02 & \second{25.05} & 0.03 \\
FT-FSOD & 16.88 & 22.80 & 19.50 & \second{12.80} & 15.90 & 25.20 &  & 21.04 & 10.31 & \second{0.12} & 13.84 & \second{0.23} \\
\ours \textbf{Struct2Det (ours)} & \best{33.01} & \best{44.17} & \best{37.65} & \best{17.63} & \best{29.69} & \best{41.78} &  & \best{36.60} & \best{37.72} & \best{0.56} & \best{37.15} & \best{1.10} \\
\mods{$\Delta$ vs.\ best baseline} & \dpos{6.09} & \dpos{7.27} & \dpos{6.65} & \dpos{4.83} & \dpos{5.49} & \dpos{9.78} &  & \dpos{4.03} & \dpos{6.02} & \dpos{0.44} & \dpos{12.10} & \dpos{0.87} \\[-2pt]
\midrule
\group{13}{Digitize-PID} \\
RegionCLIP & 10.59 & 21.03 & 9.13 & 0.43 & 10.19 & 33.04 &  & 11.69 & 17.51 & 0.00 & 14.02 & 0.00 \\
CORA & 7.97 & 25.22 & 2.92 & 0.19 & 5.65 & 21.73 &  & 11.54 & 3.97 & 0.07 & 5.90 & 0.13 \\
YOLO-World & \second{68.82} & 71.00 & 70.90 & 50.50 & \second{68.20} & 70.50 &  & \second{93.66} & \second{54.80} & 0.03 & \second{69.14} & 0.07 \\
Grounding DINO & 65.50 & 71.70 & 71.30 & \second{73.87} & 66.43 & 65.75 &  & 88.78 & 52.69 & 0.69 & 66.14 & 1.38 \\
SSDPID & 53.44 & 60.60 & 60.60 & \best{83.50} & 54.90 & 47.10 &  & 55.56 & 46.37 & \na & 50.55 & \na \\
YOLOE & 18.14 & 44.61 & 27.36 & 25.81 & 24.59 & 37.19 &  & 25.15 & 11.41 & \best{1.51} & 15.70 & \best{2.86} \\
HD-OVD & 53.92 & 64.59 & 64.47 & 34.83 & 52.68 & 66.01 &  & 78.54 & 25.77 & 0.00 & 38.81 & 0.00 \\
FT-FSOD & 66.91 & \second{71.80} & \second{71.60} & 50.10 & 66.50 & \second{72.30} &  & 92.98 & 45.82 & \second{1.08} & 61.39 & \second{2.14} \\
\ours \textbf{Struct2Det (ours)} & \best{71.08} & \best{73.80} & \best{73.40} & 58.00 & \best{71.40} & \best{73.70} &  & \best{94.37} & \best{64.48} & 0.03 & \best{76.61} & 0.07 \\
\mods{$\Delta$ vs.\ best baseline} & \dpos{2.26} & \dpos{2.00} & \dpos{1.80} & \dneg{25.50} & \dpos{3.20} & \dpos{1.40} &  & \dpos{0.71} & \dpos{9.68} & \dneg{1.48} & \dpos{7.47} & \dneg{2.79} \\[-2pt]
\bottomrule
\end{tabularx}
\end{table*}

\subsection{Benchmark and Setup}\label{sec:BAS}

InduSym is an industrial-symbol detection benchmark in which every category has a CSG specifying its components, attributes, relations, and any connection anchors. Its 674 drawings are split 539/67/68 into training, validation, and test sets before cropping, so no drawing spans two splits. Cropping with $1024\times1024$ windows at stride 768 yields 28,322 crops and 88,912 instances across 275 categories.

Categories are grouped by training-instance count into 204 base, 45 few-shot, and 26 zero-shot categories; 176 base categories have annotated test instances. Only five support instances per few-shot category are used (225 in total). Zero-shot validation annotations are not used for training, checkpoint selection, or hyperparameter tuning. Benchmark statistics, cropping, and graph construction are detailed in the Supplementary Material.

\smallskip
\noindent\textbf{Digitize-PID and Baselines.} We also evaluate on Digitize-PID~\cite{paliwal2021digitizepid}, whose 500 drawings (360/40/100 train/\allowbreak validation/\allowbreak test) yield 23,185 crops and 59,498 instances with the same cropping; its 32 categories comprise 20 base, 6 few-shot, and 6 zero-shot. We compare Struct2Det with FT-FSOD~\cite{yu2026acloser} and seven open-vocabulary or industrial-symbol detectors~\cite{zhong2022regionclip,wu2023cora,liu2024grounding,wang2025yoloe,cheng2024yoloworld,gupta2024ssdpid,fu2025hdovd} on both benchmarks. Methods share splits, support instances, and categories within each evaluated group; baselines use their official training and inference settings.

\smallskip
\noindent\textbf{Evaluation Metrics.} Both test sets use COCO AP, AP$_{50}$, AP$_{75}$, AP$_s$, AP$_m$, AP$_l$, and group-wise AP. Overall AP weights group AP by the number of annotated test categories: 176/45/26 (base/few-shot/zero-shot) on InduSym and 20/6/6 on Digitize-PID. SSDPID's overall AP excludes zero-shot categories.
\Hbf{} and \Hbz{} are harmonic means, $H(x,y)=2xy/(x+y)$, of \APb{} with \APf{} and \APz{}.

\smallskip
\noindent\textbf{Training Details.} Struct2Det uses YOLO-World and two-phase AdamW training (weight decay $0.05$); both phases enable refinement and freeze only the text encoder. Phase I trains on base categories for 12 epochs with an initial learning rate of $3.0\times10^{-5}$ (linear warm-up and decay). Phase II adds five support instances per few-shot category and trains for 3,000 iterations with a learning rate of $2.0\times10^{-5}$ (linear warm-up, cosine decay). GT-class injection (Sec.~\ref{sec:objective}) decays to zero per phase and is disabled at inference; see the Supplementary Material for training details.
All experiments use one NVIDIA A800 GPU.

\subsection{Main Results}

\begin{figure*}[t]
    \centering
    \includegraphics[width=\textwidth]{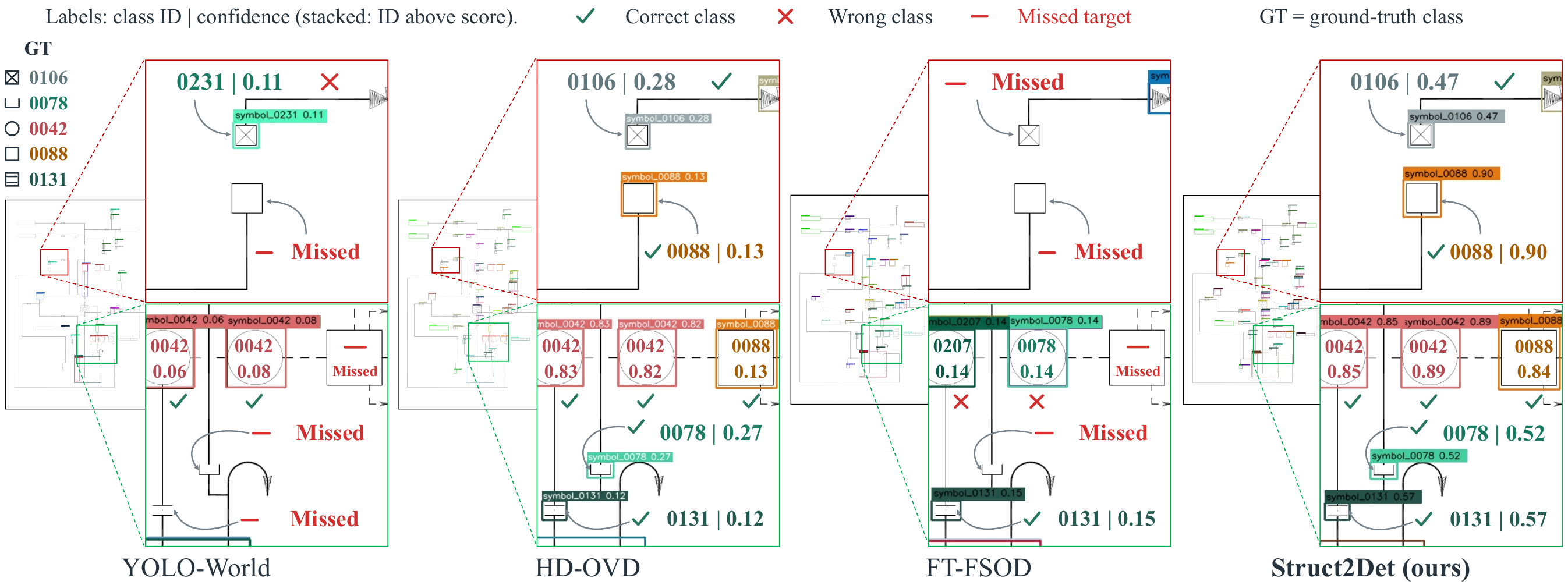}
    \caption{\textbf{Qualitative comparison on InduSym.} Two regions randomly cropped from a large engineering drawing are enlarged, highlighting seven targets. Label colors match prediction boxes.}\label{fig:qualitative_InduSym}
\end{figure*}

Table~\label{firstref:tab:main-results}\ref{tab:main-results} compares Struct2Det with eight baselines on InduSym and Digitize-PID. Struct2Det achieves 37.72\% and 64.48\% \APf{}, exceeding the strongest few-shot baseline, YOLO-World, by 6.02 and 9.68 points, respectively. It also achieves the highest \APb{} and \Hbf{} on both benchmarks.

Beyond category-wise AP, Struct2Det leads on all six detection metrics on InduSym and all except AP$_s$ on Digitize-PID, where it reaches 71.08\% overall AP. Its AP$_s$ improves by 7.50 points over YOLO-World to 58.00\%, but ranks third behind SSDPID and Grounding DINO. Limited component detail may hinder matching at small scales, although this explanation requires further analysis.

Figure~\label{firstref:fig:qualitative_InduSym}\ref{fig:qualitative_InduSym} shows misses and class confusions by YOLO-World and FT-FSOD. Both Struct2Det and HD-OVD correctly detect all seven highlighted targets, but Struct2Det assigns higher confidence to each: 0.90 vs.\ 0.13 for the upper Symbol\_0088 and 0.57 vs.\ 0.12 for Symbol\_0131. The Supplementary Material provides more examples, support-size comparisons, and computational costs.

\subsection{Is the Gain from Structure?}

Table~\label{firstref:tab:composition-ablation}\ref{tab:composition-ablation} examines structural content, image-level retrieval, and candidate-level verification, following the three stages in Sec.~\ref{sec:method}.

\begin{table}[!htb]
\fontsize{8.5}{10}\selectfont
\centering
\newcommand{\tcheck}{{\scriptsize$\checkmark$}}
\newcommand{\tmodified}{{\scriptsize$\triangle$}}
\caption{\textbf{Structure and grounding ablations on InduSym.} \tcheck: present; --: absent; \tmodified: modified CSG (text-only, wrong edges, or no edges). Refine: up to two rounds. $\Delta\mathrm{AP}_{\mathrm{few}}$: variant minus full model (points); bars show the drop magnitude.}\label{tab:composition-ablation}
\setlength{\tabcolsep}{1.3pt}
\renewcommand{\arraystretch}{1.20}
\setlength{\aboverulesep}{1.2pt}
\setlength{\belowrulesep}{1.2pt}
\newcommand{\gbar}[1]{\dbar{#1}}
\begin{tabularx}{\linewidth}{@{}X cccc @{\hspace{2.5pt}}r @{\hspace{2.5pt}}r @{\hspace{2.5pt}}r @{\hspace{3pt}}c@{}}
\toprule
\headband \tableheading{Variant} & \tableheading{\scriptsize CSG} & \tableheading{\scriptsize SGE} & \tableheading{\scriptsize MSQ} & \tableheading{\scriptsize Refine} & \APb & \APf & $H_{\text{b,f}}$ & \tableheading{\scriptsize $\Delta\mathrm{AP}_{\mathrm{few}}$} \\
\midrule
YOLO-World & -- & -- & -- & -- & 19.07 & 31.70 & 23.81 & \gbar{-6.02} \\
\midrule
Empty input & -- & \tcheck & \tcheck & \tcheck & 30.85 & 31.90 & 31.37 & \gbar{-5.82} \\
CSG as text & \tmodified & -- & -- & -- & 20.18 & 32.61 & 24.94 & \gbar{-5.11} \\
w/o relations & \tmodified & \tcheck & \tcheck & \tcheck & \second{32.75} & \second{34.35} & \second{33.53} & \gbar{-3.37} \\
Wrong relations & \tmodified & \tcheck & \tcheck & \tcheck & 32.30 & 33.16 & 32.72 & \gbar{-4.56} \\
\midrule
Global $\mathbf s_c$ & \tcheck & \tcheck & -- & \tcheck & 22.36 & 32.75 & 26.58 & \gbar{-4.97} \\
\midrule
w/o refine & \tcheck & \tcheck & \tcheck & -- & 31.42 & 32.68 & 32.04 & \gbar{-5.04} \\
\midrule
\ours \textbf{Full model} & \tcheck & \tcheck & \tcheck & \tcheck & \best{36.60} & \best{37.72} & \best{37.15} & \dbarfull \\
\bottomrule
\end{tabularx}
\end{table}

\noindent\textbf{Category level.} Empty input contains no CSG nodes or edges; checkmarks denote retained modules. It gains 11.78 base AP points but only 0.20 few-shot AP points over YOLO-World. Relative to Empty input, correct CSGs add 5.75 and 5.82 points, respectively, supporting their value for few-shot detection. CSG as text yields 32.61\% few-shot AP, below the full model's 37.72\%. w/o relations retains correct nodes and attributes but removes all edges; restoring correct edges adds 3.37 few-shot AP points. Wrong relations permutes edge endpoints while retaining node content and scores 1.19 few-shot AP points below w/o relations. Incorrect relations can thus be less useful than omitting them.

\smallskip
\noindent\textbf{Image level.} The Global $\mathbf s_c$ variant retains SGE and refinement but omits image-wide component retrieval. With correct CSGs, restoring retrieval improves base and few-shot AP by 14.24 and 4.97 points, respectively. This comparison does not isolate structural guidance from other effects of retrieval.

\smallskip
\noindent\textbf{Candidate level.} With CSG, SGE, and MSQ retained, verification and feedback improve base and few-shot AP by 5.18 and 5.04 points over w/o refine. Local correction complements image-wide retrieval; Sec.~\ref{sec:DVH} further examines NRM and LSF.

\subsection{Does Verification Help?}\label{sec:DVH}

\begin{figure}[!htb]
\centering
\begingroup
\definecolor{verifybase}{HTML}{527DA5}
\definecolor{verifyfew}{HTML}{C47761}
\definecolor{verifyink}{HTML}{48515B}
\definecolor{verifymuted}{HTML}{707B85}
\definecolor{verifygrid}{HTML}{E5E9ED}
\begin{tikzpicture}[
  y=1.45cm,
  font=\footnotesize,
  line cap=round,
  line join=round,
  every node/.style={inner sep=1.5pt, text=verifyink},
  trend/.style={line width=1.05pt},
  point/.style={line width=0.85pt},
  small/.style={font=\footnotesize},
  muted/.style={font=\scriptsize, text=verifymuted},
  tile/.style={rounded corners=1.2pt}
]
\path[use as bounding box] (-0.66,-0.85) rectangle (7.46,2.32);
\fill[verifygrid!38] (2.63,0) rectangle (4.43,1.92);
\foreach \ap in {30,32,34,36,38} {
  \draw[verifygrid, line width=0.35pt] (0,{(\ap-30)*0.24}) -- (4.43,{(\ap-30)*0.24});
  \node[anchor=east] at (-0.12,{(\ap-30)*0.24}) {\ap};
}
\draw[verifyink!65, line width=0.45pt] (0,2.06) -- (0,0) -- (4.43,0);
\foreach \r/\x in {0/0.20,1/0.95,2/1.70} {
  \draw[verifyink!65, line width=0.45pt] (\x,0) -- (\x,-0.065);
  \node[anchor=north] at (\x,-0.11) {\r};
}
\foreach \name/\x in {NRM/3.05,LSF/3.98} {
  \draw[verifyink!65, line width=0.45pt] (\x,0) -- (\x,-0.065);
  \node[small,anchor=north,align=center] at (\x,-0.115) {w/o\\\name};
}
\node[small,anchor=north] at (0.95,-0.60) {(a) Refinement rounds};

\node[anchor=west] at (-0.57,2.20) {AP (\%)};
\draw[verifybase,trend] (0.95,2.20) -- (1.33,2.20);
\draw[verifybase,fill=white,point] (1.14,2.20) circle (2.0pt);
\node[anchor=west] at (1.40,2.20) {AP$_{\text{base}}$};
\draw[verifyfew,trend] (2.88,2.20) -- (3.26,2.20);
\draw[verifyfew,fill=verifyfew,point] (3.07,2.20) circle (2.0pt);
\node[anchor=west] at (3.33,2.20) {AP$_{\text{few}}$};
\draw[verifybase,trend]
  (0.20,{(31.42-30)*0.24}) -- (0.95,{(32.83-30)*0.24}) -- (1.70,{(36.60-30)*0.24});
\draw[verifyfew,trend]
  (0.20,{(32.68-30)*0.24}) -- (0.95,{(35.29-30)*0.24}) -- (1.70,{(37.72-30)*0.24});
\foreach \x/\ap in {0.20/31.42,0.95/32.83,1.70/36.60} {
  \draw[verifybase,fill=white,point] (\x,{(\ap-30)*0.24}) circle (2.0pt);
}
\foreach \x/\ap in {0.20/32.68,0.95/35.29,1.70/37.72} {
  \draw[verifyfew,fill=verifyfew,point] (\x,{(\ap-30)*0.24}) circle (2.0pt);
}
\node[anchor=west,fill=white,inner sep=1pt,font=\footnotesize\bfseries]
  at (1.85,{(36.60-30)*0.24-0.045}) {36.60};
\node[anchor=west,fill=white,inner sep=1pt,font=\footnotesize\bfseries]
  at (1.85,{(37.72-30)*0.24+0.045}) {37.72};
\foreach \x/\few/\base in {3.05/35.02/31.79,3.98/34.87/32.08} {
  \draw[verifybase,fill=white,point] (\x,{(\base-30)*0.24}) circle (2.0pt);
  \draw[verifyfew,fill=verifyfew,point] (\x,{(\few-30)*0.24}) circle (2.0pt);
  \node[small,anchor=south,inner sep=1pt] at (\x,{(\few-30)*0.24+0.085}) {\few};
  \node[small,anchor=north,inner sep=1pt] at (\x,{(\base-30)*0.24-0.085}) {\base};
}
\node[small,anchor=south,align=center] at (5.99,1.68) {w/o\\Refine};
\node[small,anchor=south,align=center] at (6.99,1.68) {\phantom{w/o}\\Refine};
\node[small,anchor=east,align=right] at (5.46,1.25) {w/o\\MSQ};
\node[small,anchor=east] at (5.46,0.41) {MSQ};
\fill[verifygrid!90,tile] (5.52,0.86) rectangle (6.46,1.64);
\fill[verifyfew!11,tile]  (6.52,0.86) rectangle (7.46,1.64);
\fill[verifyfew!11,tile]  (5.52,0.02) rectangle (6.46,0.80);
\fill[verifyfew!45,tile]  (6.52,0.02) rectangle (7.46,0.80);
\node[font=\footnotesize] at (5.99,1.37) {31.70};
\node[muted] at (5.99,1.08) {baseline};
\node[font=\footnotesize] at (6.99,1.37) {32.75};
\node[small] at (6.99,1.08) {+1.05};
\node[font=\footnotesize] at (5.99,0.53) {32.68};
\node[small] at (5.99,0.24) {+0.98};
\node[font=\footnotesize\bfseries] at (6.99,0.53) {37.72};
\node[font=\scriptsize\bfseries] at (6.99,0.24) {+6.02};
\node[small,anchor=north] at (6.49,-0.60) {(b) AP$_{\text{few}}$ (\%)};
\end{tikzpicture}
\endgroup
\caption{\textbf{Candidate verification on InduSym.} (a) Refinement rounds and two-round ablations; open/filled: \APb{}/\APf{}. (b) \APf{} by MSQ/refinement setting; the cell with neither MSQ nor refinement is YOLO-World.}\label{fig:verification}
\end{figure}
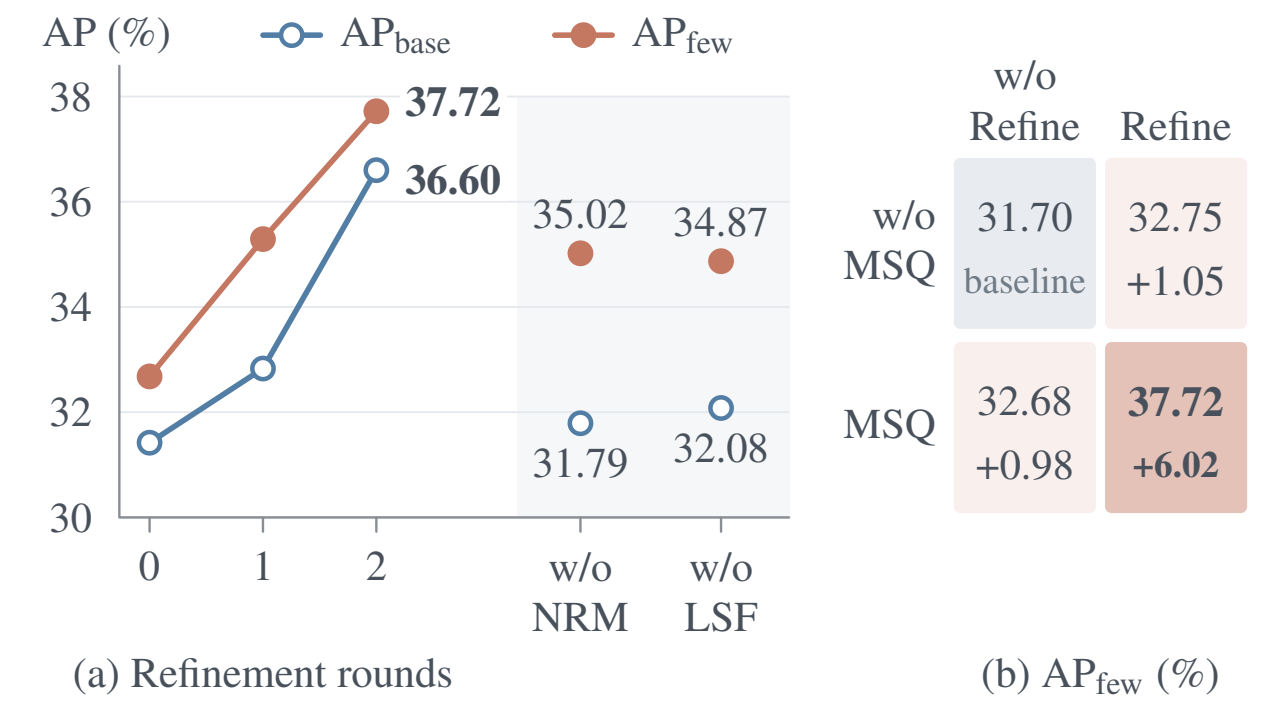

Figure~\label{firstref:fig:verification}\ref{fig:verification}(a) compares refinement rounds and variants that replace NRM or LSF. Each round helps: \APb{}/\APf{} is 31.42/32.68 with no round, 32.83/35.29 with one, and 36.60/37.72 with two.

Both variants use two rounds. w/o NRM uses the predicted class's mean component embedding without visual tokens, taking the class score as reliability. w/o LSF spatially averages and globally broadcasts each scale's residual, setting reliability maps to one and ambiguity maps to zero. w/o LSF retains NRM, SCM, projections, and clipping; spatial support, conditioning, and magnitude change jointly, so locality is not isolated. w/o NRM and w/o LSF reach 35.02\% and 34.87\% \APf{}, respectively: 2.70 and 2.85 points below the full two-round model. Both trail the full one-round model (35.29\%).

In Figure~\ref{fig:verification}(b), MSQ or refinement alone adds about one point of \APf{} over YOLO-World; together they add 6.02. As YOLO-World also lacks SGE, their interaction is not isolated.

\subsection{Beyond Engineering Drawings}

We test category structure beyond industrial symbols through natural-object detection and Chinese character recognition.

\smallskip
\noindent\textbf{(1) Natural object detection.}
We follow the OV-COCO split of OVR-CNN~\cite{zareian2021open}: 48 base and 17 novel COCO2017~\cite{lin2014coco} categories, evaluated in 5-shot and zero-shot settings. Qwen3-VL-Plus generates SVGs from category names and public definitions; the parser converts them into CSGs encoding parts, attributes, and relations.

\begin{table}[!htb]
\small
\centering
\caption{\textbf{Natural-object detection on OV-COCO} (48 base/17 novel; AP$_{50}$). All$^\dagger$: category-count-weighted mean. \na: not reported; RegionCLIP, CORA, and HD-OVD (Faster R-CNN) zero-shot results are from their papers.}\label{tab:ovcoco}
\setlength{\tabcolsep}{2.5pt}
\renewcommand{\arraystretch}{0.96}
\setlength{\aboverulesep}{1.2pt}
\setlength{\belowrulesep}{1.2pt}
\begin{tabularx}{\linewidth}{l *{3}{Y} c *{3}{Y}}
\toprule
 & \multicolumn{3}{c}{\tableheading{5-shot}} & & \multicolumn{3}{c}{\tableheading{Zero-shot}} \\
\cmidrule(lr){2-4}\cmidrule(l){6-8}
\headband \tableheading{Method} & Base & Novel & All$^\dagger$ & & Base & Novel & All$^\dagger$ \\
\midrule
RegionCLIP & \na & \na & \na & & 57.10 & 31.40 & 50.38 \\
CORA & \na & \na & \na & & 44.50 & 41.70 & 43.77 \\
YOLO-World & \second{54.72} & \second{38.14} & \second{50.38} & & \second{58.83} & \second{51.30} & \second{56.86} \\
Grounding DINO & 52.14 & 34.43 & 47.51 & & 54.32 & \best{57.05} & 55.03 \\
YOLOE & 38.32 & 36.78 & 37.92 & & 49.94 & 51.18 & 50.26 \\
HD-OVD & \na & \na & \na & & 57.70 & 46.30 & 54.72 \\
\ours \textbf{Struct2Det (ours)} & \best{55.22} & \best{39.64} & \best{51.15} & & \best{59.00} & 50.84 & \best{56.87} \\
\bottomrule
\end{tabularx}
\end{table}

In Table~\label{firstref:tab:ovcoco}\ref{tab:ovcoco}, Struct2Det leads the methods evaluated with five shots on base, novel, and overall AP$_{50}$, improving novel AP$_{50}$ by 1.50 points over YOLO-World. Under zero-shot evaluation, novel AP$_{50}$ trails YOLO-World and Grounding DINO by 0.46 and 6.21 points; base and overall AP$_{50}$ are on par with YOLO-World.

\smallskip
\noindent\textbf{(2) Character recognition.}
Characters provide a complementary test: categories share components but differ in their spatial arrangements, which Qwen3-VL-Plus converts from IDSs into structured SVGs for CSG construction. We use 1,500 \mbox{GB2312-80} level-1 classes for initial training on CASIA-HWDB~\cite{liu2011casia} and evaluate on 1,000 initially unseen classes in \mbox{ICDAR2013}~\cite{yin2013icdar}; remaining classes serve as distractors. Five labeled samples per initially unseen class support adaptation in the 5-shot setting; zero-shot uses none. We compare with SLD~\cite{chen2021zeroshotchar} and ACPM~\cite{chen2022acpm}.

\begin{table}[!htb]
\small
\centering
\caption{\textbf{Chinese character recognition on \mbox{ICDAR2013}.} Top-1 accuracy (\%) under 5-shot and zero-shot settings.}\label{tab:icdar}
\setlength{\tabcolsep}{2.5pt}
\renewcommand{\arraystretch}{0.96}
\setlength{\aboverulesep}{1.2pt}
\setlength{\belowrulesep}{1.2pt}
\begin{tabularx}{\linewidth}{l YY}
\toprule
\headband \tableheading{Method} & 5-shot & Zero-shot \\
\midrule
SLD & 71.28 & \second{22.69} \\
ACPM & \best{78.66} & 3.59 \\
\ours \textbf{Struct2Det (ours)} & \second{77.62} & \best{37.52} \\
\bottomrule
\end{tabularx}
\end{table}

In Table~\label{firstref:tab:icdar}\ref{tab:icdar}, Struct2Det reaches 77.62\% top-1 accuracy in the 5-shot setting, 1.04 points below ACPM. Under zero-shot evaluation, however, it reaches 37.52\%, exceeding the strongest baseline, SLD, by 14.83 points.

\begin{figure}[!ht]
\centering
\begingroup
\definecolor{gainours}{HTML}{527DA5}
\definecolor{gainink}{HTML}{48515B}
\definecolor{gainmuted}{HTML}{707B85}
\definecolor{gaingrid}{HTML}{E5E9ED}
\begin{tikzpicture}[
  y=0.745cm,
  font=\footnotesize,
  line cap=round,
  every node/.style={inner sep=1.5pt,text=gainink},
  muted/.style={font=\scriptsize,text=gainmuted}
]
\path[use as bounding box] (-0.05,-0.95) rectangle (7.80,5.25);
\foreach \s in {20,30,40,50,60,70} {
  \draw[gaingrid, line width=0.35pt] ({3.10+(\s-20)*0.078},-0.12) -- ({3.10+(\s-20)*0.078},4.00);
  \node[muted,anchor=north] at ({3.10+(\s-20)*0.078},-0.14) {\s};
}
\node[muted,anchor=north west] at (7.12,-0.14) {\%};
\draw[gainmuted,fill=white,line width=0.8pt] (3.20,4.90) circle (2.0pt);
\node[font=\scriptsize,anchor=west] at (3.32,4.90) {Baseline};
\fill[gainours] (5.72,4.90) circle (2.3pt);
\node[font=\scriptsize,anchor=west] at (5.84,4.90) {Struct2Det};
\newcommand{\gainrow}[5]{%
  \node[anchor=east] at (2.95,#1) {#2};
  \draw[gainours!38, line width=3.0pt, line cap=butt] ({3.10+(#3-20)*0.078},#1) -- ({3.10+(#4-20)*0.078},#1);
  \draw[gainmuted,fill=white,line width=0.9pt] ({3.10+(#3-20)*0.078},#1) circle (2.3pt);
  \draw[white,fill=gainours,line width=0.6pt] ({3.10+(#4-20)*0.078},#1) circle (3.0pt);
  \node[anchor=west,fill=white,inner sep=1pt,font=\footnotesize\bfseries] at ({3.10+(#4-20)*0.078+0.15},#1) {+#5};
}
\node[anchor=west,font=\scriptsize\bfseries] at (0.02,4.12) {5-shot detection};
\node[anchor=west,font=\scriptsize\bfseries] at (0.02,0.90) {Zero-shot recognition};
\draw[gaingrid,line width=0.5pt] (0.02,1.15) -- (7.75,1.15);
\gainrow{3.5}{\textbf{InduSym}\,{\footnotesize\color{gainmuted}AP$_{\text{few}}$}}{31.70}{37.72}{6.02}
\gainrow{2.55}{\textbf{Digitize-PID}\,{\footnotesize\color{gainmuted}AP$_{\text{few}}$}}{54.80}{64.48}{9.68}
\gainrow{1.6}{\textbf{OV-COCO}\,{\footnotesize\color{gainmuted}AP$_{50}^{\text{novel}}$}}{38.14}{39.64}{1.50}
\gainrow{0.20}{\textbf{ICDAR2013}\,{\footnotesize\color{gainmuted}top-1}}{22.69}{37.52}{14.83}
\end{tikzpicture}
\endgroup
\caption{\textbf{Gains across tasks.} Struct2Det versus the baseline (YOLO-World; SLD on ICDAR2013) in each task's metric. Detection rows are 5-shot; ICDAR2013 is zero-shot.}\label{fig:structure-gain-spectrum}
\end{figure}
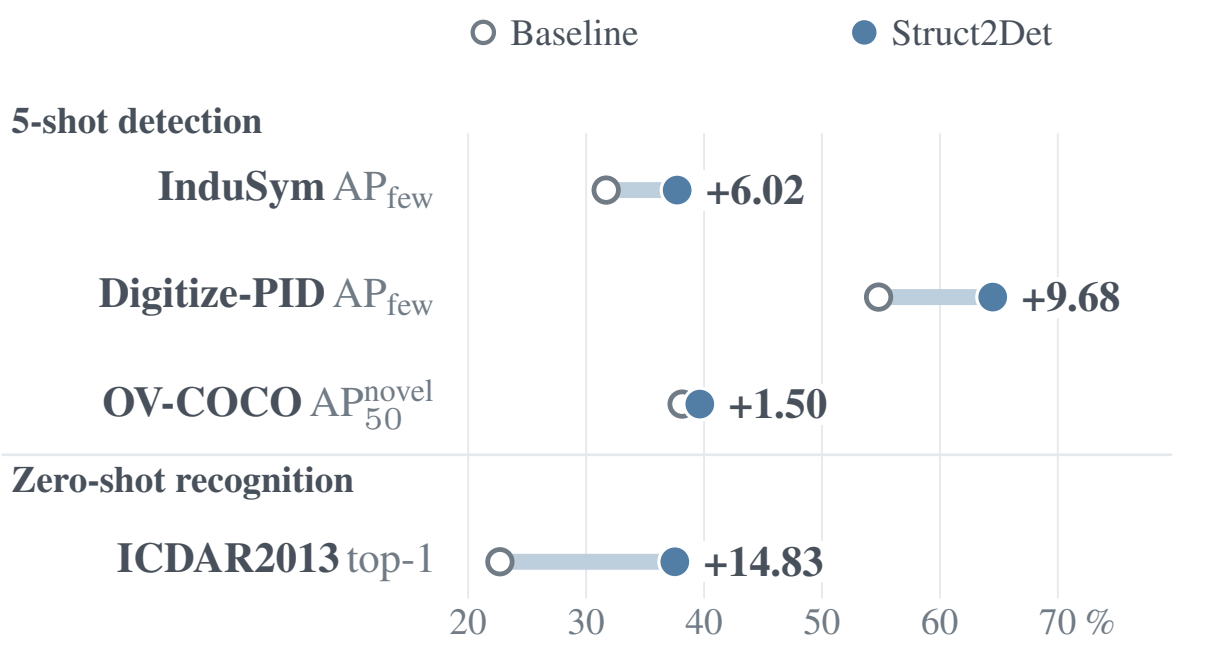

Figure~\label{firstref:fig:structure-gain-spectrum}\ref{fig:structure-gain-spectrum} suggests two patterns: structural specificity aids few-shot discrimination, while component sharing supports zero-shot transfer. Symbols' larger few-shot gains accompany stable, category-defining configurations; natural-object parts vary with viewpoint and deformation. Shared character components may let the model transfer learned visual correspondences to unseen compositions. Near-zero symbol \APz{}, however, shows that explicit structure alone is insufficient. Since tasks differ in metrics and protocols, controlled experiments are needed to isolate these effects.

\setlength{\intextsep}{10pt plus 2pt minus 2pt}

\section{Conclusion}

Struct2Det operates in three stages: encoding category structure, retrieving image-wide component evidence to modulate category text embeddings, and correcting region features through candidate-level visual--structural matching. It raises few-shot AP by 6.02 and 9.68 points on InduSym and Digitize-PID, with the highest base AP on both. On InduSym, wrong or missing relations cost 4.56 or 3.37 points of few-shot AP; removing retrieval or refinement costs 4.97 or 5.04. Further tests show few-shot gains on natural objects and zero-shot gains on Chinese characters. We will release InduSym and our code.

\smallskip
\noindent\textbf{Limitations and future work.} Zero-shot symbol detection remains weak; future work will study how to learn reliable visual--structural correspondences with less supervision. Controlled evaluations are needed to test whether limited component detail hinders small-object matching and whether shared components explain zero-shot transfer in character recognition.

\section{LLM Usage Statement}

We used OpenAI GPT-6.1 Sol for language polishing and Qwen3-VL-Plus to generate category SVGs for the datasets other than InduSym.

\section{Acknowledgments}

This work was supported in part by the Key R\&D Program of the Innovation Technology Plan of Hubei Province under Grants 2026BEB046, and in part by the Key Program of Hubei Provincial Natural Science Foundation (Xiangyang Innovation and Development Joint Fund) under Grand 2025AFD050, and in part by the National Natural Science Foundation of China under Grant 62171327,62171328 and 62072350, and in part by the Supported by Graduate Innovative Fund of Wuhan Institute of Technology under Grant CX2025289.

{
    \small
    \setlength{\baselineskip}{10.45pt}
    \setlength{\bibsep}{0.3pt}
    \bibliographystyle{ieeenat_fullname}
    \bibliography{main}
}

\addtocontents{toc}{\protect\setcounter{tocdepth}{2}}
\clearpage
\setcounter{page}{1}
\maketitlesupplementary
\appendix
\setcounter{tocdepth}{2}
\begingroup
\renewcommand{\contentsname}{Contents}
\tableofcontents
\endgroup

\setlength{\textfloatsep}{8pt plus 1pt minus 1pt}
\setlength{\dbltextfloatsep}{8pt plus 1pt minus 1pt}
\setlength{\floatsep}{8pt plus 1pt minus 1pt}
\setlength{\dblfloatsep}{8pt plus 1pt minus 1pt}
\captionsetup{skip=3pt}

\flushbottom
\setlength{\parskip}{0pt plus 2pt}

\AtBeginEnvironment{table}{\renewcommand{\arraystretch}{1.16}}

\section{Implementation Details}

The following details use the main paper's stages and notation.

\subsection{Relational Structure Encoding}

SGE uses a structural dimension of $d=256$ and $L_{\mathrm g}=4$ relation-aware message-passing layers; at layer $\tau$, node $i$ updates as
\begin{equation}
\begin{aligned}
\mathbf m_{ji}^{(\tau)}&=\phi_{\mathrm m}\!\left([\mathbf h_j^{(\tau-1)};\mathbf e_{ji}]\right),\\
\mathbf h_i^{(\tau)}&=\operatorname{LN}\!\Bigl(\mathbf h_i^{(\tau-1)}+\phi_{\mathrm u}\!\Bigl([\mathbf h_i^{(\tau-1)};\textstyle\sum_{j\to i}\mathbf m_{ji}^{(\tau)}]\Bigr)\Bigr),
\end{aligned}
\end{equation}
where $[\cdot;\cdot]$ concatenates, $\mathbf e_{ji}$ is the relation embedding, and $\phi_{\mathrm m},\phi_{\mathrm u}$ are two-layer MLPs. For the visual-component nodes $i\in\mathcal V_c^{+}$ of category $c$ (all nodes except roots and groups), $\mathbf h_{c,i}^{+}=\mathbf h_i^{(L_{\mathrm g})}$. The three readouts are computed as
\begin{equation}
\begin{gathered}
\mathbf g_c=\textstyle\sum\nolimits_{i}\pi_i\,\mathbf h_{c,i}^{+},\qquad
\mathbf r_c=\max\nolimits_{i}\mathbf h_{c,i}^{+},\\[2pt]
\mathbf p_c=\tfrac{1}{P}\textstyle\sum\nolimits_{\nu=1}^{P}\sum\nolimits_{i}\mathrm{softmax}\!\left(\tfrac{\mathbf q_\nu^{\top}\mathbf h_{c,i}^{+}}{\sqrt{d}}\right)\mathbf h_{c,i}^{+}
\end{gathered}
\end{equation}
where $\pi_i$ are learned salience weights, softmax-normalized over up to eight salient nodes, and $\mathbf q_\nu$ are $P=8$ learned query probes. The maximum in $\mathbf r_c$ is element-wise, and the softmax in $\mathbf p_c$ is over component nodes. The three readouts are concatenated and projected by the two-layer MLP $\phi_{\mathrm o}$ to the global condition $\mathbf s_c\in\mathbb R^{256}$.

\subsection{Component-Grounded Detection}

\noindent\textbf{Multi-scale sampling.} MSQ uses $N_{\mathrm h}=8$ heads and $K=4$ samples per component, head, and feature level, with $\ell\in\{3,4,5\}$. Two linear heads predict offsets and reading weights from the component embedding; weights are softmax-normalized over the $12$ level--point cells within each head. A component embedding predicts an image-independent reference point $\mathbf o_i=\sigma(\mathbf W_{\mathrm{ref}}\mathbf h_{c,i}^{+})$ in normalized image coordinates. For feature width $W_\ell$ and height $H_\ell$, the sampling location is
\begin{equation}
\boldsymbol\xi_{i,h,\ell,u}=\mathbf o_i+\boldsymbol\Delta_{i,h,\ell,u}\odot(W_\ell^{-1},H_\ell^{-1}),
\end{equation}
where $h$ indexes heads, $u$ indexes samples, and $\boldsymbol\Delta_{i,h,\ell,u}$ is the predicted offset. Locations are clamped to $[-0.5,1.5]^2$; samples outside the image read zero.

\noindent\textbf{MSQ read-out.} Let $\mathbf v_i$ denote the feature retrieved for component $i$ and $\hat A_{ij}$ the row-normalized graph adjacency. One residual smoothing step and mean pooling produce the raw image evidence:
\begin{equation}
\begin{aligned}
\mathbf v_i&\leftarrow\operatorname{LN}\!\Bigl(\mathbf v_i+\sum_{j}\hat A_{ij}\mathbf v_j\Bigr),\\
\mathbf z_{n,c}&=\operatorname{LN}\!\Bigl(\frac{1}{|\mathcal V_c^{+}|}\sum_{i\in\mathcal V_c^{+}}\mathbf v_i\Bigr).
\end{aligned}
\end{equation}

\noindent\textbf{Condition fusion.} Within MSQ, $\hat{\mathbf z}_{n,c}=\operatorname{LN}(\mathbf s_c+\alpha\operatorname{proj}(\mathbf z_{n,c}))$, where the nonnegative learnable $\alpha$ is initialized to zero. This combines the retrieved evidence with the global condition before text modulation.

The text embedding and fused MSQ output are first projected to a $512$-d fusion space ($\mathbf P_t\!:\!512\!\to\!512$, $\mathbf P_s\!:\!256\!\to\!512$), with projected inputs denoted by $\mathbf t'$ and $\hat{\mathbf z}'$. The fusion gates them element-wise and adds a residual derived from their mixture and difference:
\begin{equation}
\begin{gathered}
\boldsymbol\kappa=\sigma\!\left(\mathbf W_g[\mathbf t';\hat{\mathbf z}']\right),\quad
\boldsymbol\chi=\boldsymbol\kappa\odot\mathbf t'+(\mathbf 1-\boldsymbol\kappa)\odot\hat{\mathbf z}',\\
\phi_{\mathrm{fuse}}(\mathbf t,\hat{\mathbf z})=\mathbf t'+\gamma\,\phi_g\!\left([\boldsymbol\chi;\mathbf t';\mathbf t'-\hat{\mathbf z}']\right),
\end{gathered}
\end{equation}
with inner weight $\gamma=0.1$ and projections $\mathbf W_g\!:\!1024\!\to\!512$, $\phi_g\!:\!1536\!\to\!512$; the two-layer adapter $\psi$ ($512\to1024\to512$, LayerNorm, L2) and the outer residual give $\widetilde{\mathbf t}=\operatorname{LN}(\mathbf t+\lambda\,\psi(\phi_{\mathrm{fuse}}(\mathbf t,\hat{\mathbf z})))$ with $\lambda=0.08$. The inner residual preserves the projected text embedding $\mathbf t'$; the outer one adds the adapted fusion result to the original $\mathbf t$.

\subsection{Verification-Guided Refinement}

\noindent\textbf{Candidate matching and filtering.} NRM matches the top $128$ candidates against five category hypotheses each. It samples a $3\times3$ grid by bilinear interpolation at fractions $\{1/6,1/2,5/6\}$ along each box dimension. Visual tokens and component embeddings share a $128$-dimensional space.
For each candidate and category, NRM selects the top four grid--node pairs by raw score, reweights the scores by $0.75+0.25b$ ($b=1$ for mutual nearest matches and $0$ otherwise), and discards scores below $0.55$. Greedy filtering follows the original ranking and does not refill rejected slots. Pairs conflict if both grid coordinates differ by at most one cell, or if their nodes coincide or share an edge in either direction. Sibling edges limit retention to one component per parent and hypothesis. Inference stops when no candidate passes the gate or after $R_{\max}=2$ rounds; training follows Algorithm~\ref{alg:struct2det-training}.

\noindent\textbf{Correction aggregation.} For each retained pair, let $\mathbf f$ and $\mathbf h$ denote the visual and component embeddings, and $\bar s$ the adjusted match score. Each of eight expert bases maps $\tfrac12(\mathbf f+\mathbf h)$ through a $128\to8\to128$ bottleneck with GELU between the projections. A separate hypernetwork predicts the mixture weights from the $385$-dimensional descriptor $[\mathbf f;\mathbf h;\mathbf f\odot\mathbf h;\bar s]$. It applies LayerNorm, a $385\to128$ projection, GELU, a $128\to8$ projection, and softmax. The weighted sum of the basis outputs passes through LayerNorm and $0.1\tanh(\cdot)$.
Within each hypothesis, pair corrections are averaged with weights $\bar s$. Across hypotheses, the softmax uses the log detector score plus $0.5$ times the strongest retained adjusted score. Hypotheses without a retained pair receive zero weight. If no pair survives across hypotheses, the candidate's correction and reliability are zero.

\SetAlFnt{\footnotesize}
\begin{algorithm}[t]
\caption{Struct2Det Training}\label{alg:struct2det-training}
\KwInput{Image $I_n$, GT, graph $G_c$, text $\mathbf t_{n,c}$,
step $s\in\{1,2,3\}$, and learning rate $\eta_{\mathrm{lr}}$.
Use $\pi$ and $\omega_r$ from the training schedule.}
\KwOutput{Updated parameters $\boldsymbol\theta$.}

$(\mathbf H_c^+,\mathbf s_c)\leftarrow\op{SGE}(G_c)$;
$\mathbf Y\leftarrow\op{Backbone}(I_n)$\;
Obtain $\hat{\mathbf z}_{n,c}$ from
$\op{MSQ}(\mathbf Y,\mathbf H_c^+,\mathbf s_c)$ and set
$\widetilde{\mathbf t}_{n,c}\leftarrow
\op{TextFusion}(\mathbf t_{n,c},\hat{\mathbf z}_{n,c})$\;
$\loss{cond}\leftarrow\op{CondLoss}
((\mathbf s_c,\widetilde{\mathbf t}_{n,c}),
 (\bar{\mathbf s}_c,\mathbf t_{n,c}))$\;

Write $P^{(r)}=(\mathbf B^{(r)},\mathbf Q^{(r)})$ for the predictions\;
$(\mathbf F^{(0)},P^{(0)})\leftarrow
\op{Detector}(\mathbf Y,\widetilde{\mathbf t}_{n,c})$\;
$\loss{det}\leftarrow\op{DetLoss}(P^{(0)},\mathrm{GT})$\;

$R_{\mathrm{sched}}\leftarrow 1$ if $s=1$, else $R_{\max}\,(=2)$\;
$R\leftarrow0$; $\Gamma_k^{(0)}\leftarrow1$ for all $k$\;
\For{$r=1,\ldots,R_{\mathrm{sched}}$}{
  $\mathcal C_k\leftarrow
  \op{CurriculumTop5}(\mathbf Q_k^{(r-1)},\mathrm{GT},\pi)$\;
  $a_k\leftarrow\Gamma_k^{(r-1)}
  \mathbf1[\exists c\in\mathcal C_k:\,|\mathcal V_c^+|>0]$\;
  \lIf{$r=1$ and $a_k=0$ for all $k$}{\Break}

  $(\bar S,\Delta\mathbf u,\rho)\leftarrow
  \op{NRM}_{\mathrm{Match}}
  (\mathbf F^{(r-1)},P^{(r-1)},\mathbf H_c^+,\mathcal C,a)$\;
  $R\leftarrow r$;
  $\loss{loc}^{(r)}\leftarrow\op{LocalLoss}(\bar S,\mathrm{GT},a)$\;

  \eIf{$r=1$ and $s\in\{1,2\}$}{
    $\Gamma_k^{(r)}\leftarrow a_k\mathbf1[\rho_k>0]$
  }{
    $\Gamma_k^{(r)}\leftarrow a_k
    \op{NRM}_{\mathrm{gate}}(\rho_k,\mathbf Q_k^{(r-1)},r)$\;
  }
  \lIf{$\Gamma_k^{(r)}=0$ for all $k$}{
    $\loss{fb}^{(r)}\leftarrow0$; \Break}

  $(\mathbf F^{(r)},\zeta)\leftarrow
  \op{LSF}(\mathbf F^{(r-1)},\mathbf B^{(r-1)},
           \Delta\mathbf u,\Gamma^{(r)})$\;
  $\loss{fb}^{(r)}\leftarrow\op{FeedbackLoss}(\zeta)$\;
  $P^{(r)}\leftarrow\op{DetectAndMerge}
  (\mathbf F^{(r)},\widetilde{\mathbf t}_{n,c},
   P^{(r-1)},\Gamma^{(r)})$\;
  $\loss{det}\leftarrow\loss{det}
  +\omega_r\op{DetLoss}(P^{(r)},\mathrm{GT})$\;
}
$\mathcal L\leftarrow\op{TotalLoss}
(\loss{det},\loss{cond},
 \{\loss{loc}^{(r)},\loss{fb}^{(r)}\}_{r=1}^{R},R)$\;
Update $\boldsymbol\theta$ with AdamW at rate $\eta_{\mathrm{lr}}$
on $\mathcal L+\mathcal L_{\mathrm{aux}}+\mathcal L_{\mathrm{backbone}}$\;
\end{algorithm}

\noindent\textbf{Local structural feedback.} At refinement round $r$, LSF maps the center of the candidate's previous-round predicted box $\mathbf B_k^{(r-1)}$ to a cell on the source pyramid level. It scatters the accepted correction over a $3\times3$ neighborhood centered on that cell using the kernel $(1,2,1)^\top(1,2,1)/16$. After reliability-weighted aggregation, the learned $1\times1$ projection $\mathbf W_\ell$ maps the $128$-dimensional residual to $C_\ell$ channels. The same neighborhoods define two maps: $\mathbf S^\rho$ averages $\rho_k^{(r)}$ with kernel weights, and $\mathbf S^m$ averages $1-m_k^{(r)}$ with kernel weights multiplied by $\rho_k^{(r)}$. Both maps are zero outside these neighborhoods.

SCM combines the spatial and channel gates as
\begin{equation}
\begin{aligned}
\mathbf M_{n,\ell}^{(r)}
&=\sigma\!\left(\phi_{\mathrm{sp}}([\mathbf F;\Delta\mathbf F;\mathbf S^{\rho};\mathbf S^{m}])\right)\\
&\quad\odot\sigma\!\left(\phi_{\mathrm{ch}}(\operatorname{GAP}[\mathbf F;\Delta\mathbf F])\right),
\end{aligned}
\end{equation}
where $\mathbf F=\mathbf F_{n,\ell}^{(r-1)}$, $\Delta\mathbf F=\Delta\mathbf F_{n,\ell}^{(r)}$, and $\operatorname{GAP}$ denotes global average pooling. The spatial gate uses $3\times3$ and $1\times1$ convolutions, and the channel gate uses two $1\times1$ convolutions. Both use GELU and hidden width $\max(16,\lfloor C_\ell/4\rfloor)$. Their sigmoid outputs broadcast over channels and spatial locations, respectively.

The update before clipping is $\mathbf D=\eta_\ell\mathbf M_{n,\ell}^{(r)}\odot\mathbf P_\ell(\Delta\mathbf F)$, where $\mathbf P_\ell$ is a channel-preserving $1\times1$ convolution. The per-level scale $\eta_\ell$ starts at $0.01$ and satisfies $|\eta_\ell|<0.1$. With zero-initialized $\mathbf P_\ell$, the initial update is zero. In FP32, the feedback path computes $\zeta_{n,\ell}^{(r)}=\operatorname{RMS}_\epsilon(\mathbf D)/\operatorname{RMS}_\epsilon(\mathbf F)$ per image and level, where $\operatorname{RMS}_\epsilon(\mathbf X)=\sqrt{\operatorname{mean}(\mathbf X^2)+10^{-12}}$. LSF scales $\mathbf D$ by $\min(1,\beta/\max(\zeta_{n,\ell}^{(r)},10^{-6}))$, with $\beta=0.10$.

Verification uses retained matches as local component evidence for each class hypothesis. A strong match to a shared component can support competing hypotheses even when a distinguishing component remains unmatched. Graph relations inform component embeddings and redundancy filtering.

\section{Training Procedure}

\noindent\textbf{Common settings.} Both phases train on $640\times640$ inputs with automatic mixed precision. The text encoder is frozen, and all other trainable components are optimized jointly. We apply random affine transformations, HSV perturbation, and horizontal flipping with probability $0.5$.

\noindent\textbf{Phase I.} The detector is trained on base categories for 12 epochs with batch size $24$. Mosaic and MixUp are additionally applied with probability $0.15$, and Mosaic is disabled during the final three epochs.

\noindent\textbf{Phase II.} For each few-shot category, training uses a support set of five instances. The stage runs for 3,000 iterations with batch size $4$, and each cycle contains one base-category batch and three few-shot batches.

\noindent\textbf{Refinement schedule.} Each phase follows three training steps $s\in\{1,2,3\}$, covering epochs 1--2, 3--4, and 5--12 in Phase I and iterations 1--500, 501--1,000, and 1,001--3,000 in Phase II. The detection-loss weights are $(\omega_1,\omega_2)=(0.1,0)$, $(0.25,0.1)$, and $(0.5,0.25)$. The loop bound is $R_{\mathrm{sched}}=1$ for step 1 and $R_{\mathrm{sched}}=R_{\max}=2$ for steps 2--3.

\noindent\textbf{Gate curriculum.} In round 1 of steps 1 and 2, NRM bypasses the candidate-level ambiguity and reliability thresholds. An active candidate passes if $\rho_k^{(r)}>0$, which requires a retained pair after link selection and conflict filtering. Let $m_k^{(r)}$ be the margin between the top two category scores before round $r$. In other training rounds and at inference, the gate requires $m_k^{(r)}<\delta_m^{(r)}$ and $\rho_k^{(r)}\ge\delta_\rho^{(r)}$, with $(\delta_m^{(1)},\delta_\rho^{(1)})=(0.15,0.55)$ for round 1 and $(\delta_m^{(2)},\delta_\rho^{(2)})=(0.10,0.60)$ for round 2. LSF uses this margin and its spatial--channel gates when applying accepted corrections throughout training.

\noindent\textbf{GT-class curriculum.} The detector's top-five hypotheses may omit the ground-truth category, preventing positive structural matching for that category. For GT-assigned training candidates missing $c^*$, \textsc{CurriculumTop5} injects $c^*$ into a fraction $\pi$ of eligible hypothesis lists, where $\pi=0.5$, $0.25$, and $0$ in steps 1--3, respectively. Injection replaces the lowest-scoring hypothesis and re-sorts the list using the original detector scores. At inference, $\pi=0$, and no ground-truth information is used.

Algorithm~\label{firstref:alg:struct2det-training}\ref{alg:struct2det-training} uses the cached category reference $\bar{\mathbf s}_c$. Candidate $k$ becomes active if at least one current hypothesis contains component nodes and, after round 1, the candidate was accepted in the preceding round. If no candidate is initially active, refinement is skipped and $R=0$. \textsc{DetectAndMerge} updates only accepted candidates, leaving other predictions unchanged, and each refinement round performs matching within the boxes produced by the preceding round. The auxiliary and backbone objectives are included in optimization.

\section{Loss Definitions}

\noindent\textbf{Overall objective.} The overall target combines the final detection objective with conditioning and round-dependent structural supervision:
\begin{equation}
\begin{aligned}
\mathcal L
={}&\loss{det}+\loss{cond}\\
&+\frac{1}{\max(1,R)}\sum_{r=1}^{R}
\left(\loss{loc}^{(r)}+\lambda_{\mathrm{fb}}\loss{fb}^{(r)}\right).
\end{aligned}
\end{equation}
Accepted refinement rounds re-predict boxes and category scores, so their detection supervision is accumulated as
\begin{equation}
\loss{det}
=
\loss{det}^{(0)}
+
\sum_{r=1}^{R}\omega_r\loss{det}^{(r)},
\end{equation}
where $\loss{det}^{(r)}$ contains classification and box-regression terms.
Each $\omega_r$ weights only the corresponding detection loss. We average the local and feedback losses over the $R\le R_{\mathrm{sched}}$ rounds that execute matching. If no candidate is accepted after matching, the round retains $\loss{loc}^{(r)}$, contributes zero feedback and detection losses, and terminates refinement. Empty index sets contribute zero.

\noindent\textbf{Conditioning consistency.} The global condition $\mathbf s_c$ is anchored to a detached, cached per-class reference $\bar{\mathbf s}_c$, while the modulated text embedding is kept close to the original embedding:
\begin{equation}
\begin{aligned}
\mathcal{L}_{\mathrm{cond}}
={}&\,\lambda_{\mathrm{cos}}\left(1-\cos(\mathbf s_c,\bar{\mathbf s}_c)\right)\\
&+\lambda_{\mathrm{L1}}\,\operatorname{SmoothL1}\!\left(\mathbf s_c,\bar{\mathbf s}_c\right)\\
&+\operatorname{ramp}(p)\lambda_{\mathrm{txt}}
\left(1-\cos(\widetilde{\mathbf t}_{n,c},\mathbf t_{n,c})\right)\\
&+\operatorname{ramp}(p)\lambda_{\mathrm{rel}}\mathcal L_{\mathrm{rel}}.
\end{aligned}
\end{equation}
Here $p\in[0,1]$ is normalized progress within the current training phase, with epoch progress in Phase~I and iteration progress in Phase~II, and $\operatorname{ramp}(p)=\min\{1,\max(0,6p-1)\}$. It is zero in step~1, increases linearly in step~2, and equals one in step~3. Let $\mathbf T$ and $\widetilde{\mathbf T}$ stack the row-normalized original and modulated text embeddings for valid image--category pairs. The relative term $\mathcal L_{\mathrm{rel}}$ is the mean SmoothL1 difference between their off-diagonal similarity matrices, and is zero when fewer than two pairs are valid. Each conditioning term is averaged over valid pairs.

\noindent\textbf{Structural supervision.} For an active candidate $k$ whose category hypotheses contain the ground-truth category $c^*$, let $s_k^{\mathrm{pos}}$ denote its strongest adjusted node match to that category and $s_k^{\mathrm{neg}}$ its strongest match among the remaining hypotheses. Both scores are measured before thresholding and conflict filtering. The local correspondence loss raises $s_k^{\mathrm{pos}}$ and requires it to exceed $s_k^{\mathrm{neg}}$ by a margin $\mu$:
\begin{equation}
\begin{aligned}
\mathcal{L}_{\mathrm{loc}}^{(r)}
&=\frac{1}{|\mathcal P^{(r)}|}\sum_{k\in\mathcal P^{(r)}}\\
&\quad\times\Bigl(\lambda_{\mathrm{pull}}\,(1-s_k^{\mathrm{pos}})
+\lambda_{\mathrm{mar}}\,[\mu-s_k^{\mathrm{pos}}+s_k^{\mathrm{neg}}]_{+}\Bigr)
\end{aligned}
\end{equation}
where $\mathcal P^{(r)}=\{k:a_k^{(r)}=1,\ c^*\in\mathcal C_k^{(r)}\}$, $[x]_+=\max(x,0)$, and $s_k^{\mathrm{neg}}=0$ when no competing hypothesis has a valid match.
For the feedback term, let $\zeta_{n,\ell}^{(r)}$ be the update-to-feature RMS ratio before clipping, defined under Local structural feedback above. With $\mathcal U^{(r)}=\{(n,\ell):\zeta_{n,\ell}^{(r)}>\beta\}$ for violating image--scale pairs in a batch, the loss is
\begin{equation}
\mathcal{L}_{\mathrm{fb}}^{(r)}
=\frac{1}{|\mathcal U^{(r)}|}\sum_{(n,\ell)\in\mathcal U^{(r)}}\left(\zeta_{n,\ell}^{(r)}-\beta\right).
\end{equation}
Measuring before clipping preserves a gradient when clipping limits the applied update. All empty terms contribute zero; $\lambda_{\mathrm{fb}}$ is applied in the overall objective above.

The coefficients are $\lambda_{\mathrm{cos}}=\lambda_{\mathrm{L1}}=0.03$, $\lambda_{\mathrm{txt}}=0.02$, $\lambda_{\mathrm{rel}}=0.005$, $\lambda_{\mathrm{pull}}=0.02$, $\lambda_{\mathrm{mar}}=0.30$, $\lambda_{\mathrm{fb}}=0.01$, $\mu=0.10$, and $\beta=0.10$.

\newcounter{indusympromptbox}
\begin{figure*}[!t]
\centering
\begingroup
\setlength{\fboxsep}{5pt}
\setlength{\fboxrule}{0.5pt}
\fbox{%
\begin{minipage}{\dimexpr\textwidth-2\fboxsep-2\fboxrule\relax}
\footnotesize
\setlength{\smallskipamount}{1pt plus 1pt minus 0.5pt}

\fontencoding{T1}\selectfont
\setlength{\JustifyingParindent}{0pt}
\justifying
\textbf{Role:} You are an engineering drawing expert familiar with industrial symbols, P\&ID legends, and SVG vector graphics.

\smallskip
\textbf{Task:} Construct a representative two-dimensional engineering symbol from the supplied category information. Preserve the specified shapes, components, and connections, and include only components supported by the input.

\smallskip
\textbf{Input:}\par
Category name: \{category name\}\par
Category definition: \{category definition\}\par
Legend or structural description: \{reference description\}; use \{None\} if unavailable.

\smallskip
\textbf{Drawing Requirements:}\par
\smallskip
\begin{minipage}[t]{0.485\linewidth}
\justifying
\textbf{1. Symbol content.} Draw only the symbol, without drawing borders, title blocks, coordinate grids, backgrounds, shadows, or unrelated decorations. Center the symbol and preserve the relative positions and proportions of its components.

\smallskip
\textbf{2. Drawing style.} Use a clear monochrome engineering drawing style with black strokes. Use white or no fill for closed regions and no fill for open lines. Avoid gradients, textures, colors, and raster images.

\smallskip
\textbf{3. Canvas.} Define an explicit SVG viewBox enclosing the entire symbol, and set width and height to match its dimensions. Leave a small, even margin around the symbol.

\smallskip
\textbf{4. Geometry and grouping.} Use only path, rect, circle, ellipse, line, polygon, and polyline for geometry. Organize primitives belonging to the same component using nested g elements.

\smallskip
\textbf{5. Identifiers and hierarchy.} Assign identifiers to the root, groups, primitives, and anchors. Maintain a clear component hierarchy. Keep distinct components separate rather than merging the entire symbol into one path, unless the symbol is intrinsically indivisible.
\end{minipage}\hfill%
\begin{minipage}[t]{0.485\linewidth}
\justifying
\textbf{6. Connection anchors.} For each pipe port, electrical terminal, or other connection location explicitly specified in the input, add an invisible circle at that location. Assign a unique id, set \mbox{class="connect"}, and name the anchors sequentially using \mbox{pointname="cp1"}, \mbox{"cp2"}, and so on. Use \mbox{r="0.5"} and \mbox{style="fill-opacity:0;}\allowbreak\mbox{stroke-width:0;}\allowbreak\mbox{stroke-opacity:0"}.

\smallskip
\textbf{7. Attachment anchors.} Use \mbox{class="adsorb"} only for adsorption or attachment locations explicitly supported by the input. Do not interpret ordinary geometric intersections as connection points.

\smallskip
\textbf{8. Element references.} When the input specifies a reference between two elements, use the target attribute to point to the referenced element's existing id. Omit this attribute when no explicit reference is provided.

\smallskip
\textbf{9. Text.} Retain letters or numbers only when they are part of the symbol itself. Do not add the category name, explanatory text, or automatically generated labels.

\smallskip
\textbf{10. Output.} Return only one complete, well-formed standalone SVG document. Do not include Markdown fences, explanations, comments, scripts, external resources, CSS stylesheets, or SVG filters. Check SVG syntax, identifier uniqueness, coordinates, and the viewBox before returning the output.
\end{minipage}
\par
\end{minipage}%
}
\par\smallskip
\refstepcounter{indusympromptbox}\label{box:indusym_prompt}%
\parbox{\textwidth}{\small\centering\textbf{Box~\theindusympromptbox.} Prompt for generating structured symbol SVGs for Digitize-PID categories.\par}
\endgroup
\end{figure*}

\noindent\textbf{Auxiliary regularization.} The auxiliary terms stabilize gate activation, candidate prediction, expert selection, and visual--structural alignment. For round $r$, let $\bar g^{(r)}$ denote the mean spatial--channel gate response, and let $q_{k,\ell,c}^{(r)}$ and $\hat q_{k,\ell,c}^{(r)}$ denote post- and detached pre-update sigmoid category probabilities. The set $\mathcal S_\ell^{(r)}$ contains accepted candidates with pre-update top score at least $0.5$ and top-two margin at least $0.15$; $\mathcal H^{(r)}$ contains levels with nonempty $\mathcal S_\ell^{(r)}$. Round consistency can therefore contribute only in the first refinement round of training steps 1 and 2, when the ambiguity threshold is bypassed. In all other training rounds, the gate accepts only candidates with a pre-update margin below $0.15$, so the consistency set is empty. Let $\hat{\mathbf e}_i$ denote normalized coefficients of selected experts, and let $\mathbf v_k$ and $\mathbf z_k$ denote row-normalized visual and structure descriptors. With $\mathcal G^{(r)}$ containing accepted candidates with valid structures, the regularizers are
\begin{equation}
\begin{aligned}
\mathcal L_{\mathrm{gate}}^{(r)}&=\left[0.10-\bar g^{(r)}\right]_+,\\
\mathcal L_{\mathrm{round}}^{(r)}&=\frac{1}{|\mathcal H^{(r)}|}\sum_{\ell\in\mathcal H^{(r)}}\frac{1}{|\mathcal S_\ell^{(r)}|C}\\
&\quad\times\sum_{k\in\mathcal S_\ell^{(r)}}\sum_{c}\operatorname{SmoothL1}\!\left(q_{k,\ell,c}^{(r)},\hat q_{k,\ell,c}^{(r)}\right),\\
\mathcal L_{\mathrm{div}}^{(r)}&=\frac{\sum_{i<j}\cos^2(\hat{\mathbf e}_i,\hat{\mathbf e}_j)}{\max\left(1,\sum_{i<j}1\right)},\\
\mathcal L_{\mathrm{roi}}^{(r)}&=\frac{1}{|\mathcal G^{(r)}|}\sum_{k\in\mathcal G^{(r)}}\left(1-\mathbf v_k^\top\mathbf z_k\right),
\end{aligned}
\end{equation}
where $C$ is the category count and $i<j$ ranges over selected expert pairs. Empty sets contribute zero, and round consistency uses level-balanced averaging. The auxiliary objective is
\begin{equation}
\begin{aligned}
\mathcal L_{\mathrm{aux}}={}&\frac{1}{\max(1,R)}\sum_{r=1}^{R}
\left(0.005\mathcal L_{\mathrm{gate}}^{(r)}\right.\\
&\left.+0.01\mathcal L_{\mathrm{round}}^{(r)}
+0.005\mathcal L_{\mathrm{div}}^{(r)}
+0.02\mathcal L_{\mathrm{roi}}^{(r)}\right).
\end{aligned}
\end{equation}

\noindent\textbf{Training-specific objectives.} The structure-aware masked branch augments YOLO-World with MAE and structural MAE reconstruction, teacher anchoring, and backbone ROI alignment, distinct from $\mathcal L_{\mathrm{roi}}$. It masks $15\%$/$30\%$ of patches and protects $25\%$ of structure tokens. During Phase~I, the teacher weight decreases from $0.05$ to $0.015$ at epoch~3 and to zero at epoch~6. These training-only terms form $\mathcal L_{\mathrm{backbone}}$ and are excluded from inference. Algorithm~\ref{alg:struct2det-training} differentiates
\begin{equation}
\mathcal L_{\mathrm{opt}}=\mathcal L+\mathcal L_{\mathrm{aux}}+\mathcal L_{\mathrm{backbone}},
\end{equation}
with the frozen text encoder excluded from the update.

\section{Structural Graph Construction}\label{sec:indusym-suppl}

We construct category structural graphs (CSGs) using a unified SVG-to-CSG pipeline across all domains. For InduSym, category SVGs are provided and the parsed graphs are verified by domain experts against their legends. For other datasets, Qwen3-VL-Plus generates semantic SVGs from category information under a shared schema. The schema standardizes coordinate systems, node types, domain attributes, and relation encodings. Box~\label{firstref:box:indusym_prompt}\ref{box:indusym_prompt} shows the prompt used for Digitize-PID, where the input consists of category names, definitions, and available legend or primitive-level descriptions. Table~\ref{tab:csg-instances} presents representative CSGs across four domains.

\begin{table*}[!t]
\centering
\caption{\textbf{Representative category structural graphs across four domains.}}
\label{tab:csg-instances}
\footnotesize
\renewcommand{\arraystretch}{1.12}
\setlength{\tabcolsep}{2pt}
\begin{fitwidthtabular}{@{}l cccc@{}}
\toprule
\multicolumn{1}{@{}>{\columncolor{headrow}[0pt][\tabcolsep]}l}{\tableheading{Example}}
& \cellcolor{headrow}$\mathcal V_c$ & \cellcolor{headrow}$\mathcal E_c$ & \cellcolor{headrow}$\mathbf{X}_c$
& \multicolumn{1}{>{\columncolor{headrow}[\tabcolsep][0pt]}c@{}}{$\mathbf{R}_c$} \\
\midrule
InduSym: NC Pneumatic Actuator with Positioner
& \makecell{16 nodes\\1 root, 1 group,\\14 primitives}
& \makecell{212 directed edges\\15 nesting and 91 sibling\\pairs, both bidirectional}
& \makecell{Primitive geometry\\and normalized boxes}
& \makecell{parent\_child,\\sibling} \\

\addlinespace[2pt]

Digitize-PID: symbol\_31
& \makecell{15 nodes\\1 root, 5 groups,\\5 primitives, 2 anchors,\\2 text nodes}
& \makecell{66 directed edges\\28 parent\_child, 34 sibling,\\4 anchor\_ref}
& \makecell{Primitive geometry and\\normalized boxes}
& \makecell{parent\_child, anchor\_ref,\\sibling} \\

\addlinespace[2pt]

OV-COCO: airplane
& \makecell{8 nodes\\1 object, 5 parts,\\2 attributes}
& \makecell{11 directed edges\\5 has\_part, 2 has\_property,\\3 attached\_to, 1 contained\_in}
& \makecell{Part role, shape,\\position and scale}
& \makecell{has\_part, has\_property,\\attached\_to, contained\_in} \\

\addlinespace[2pt]

ICDAR2013: Chinese character \cjkchar{啊}
& \makecell{8 nodes\\1 character, 3 operators,\\4 parts}
& \makecell{13 directed edges\\7 parent\_child, 2 left\_of,\\2 right\_of, 1 encloses, 1 inside}
& \makecell{Radical, normalized box,\\semantic vector}
& \makecell{parent\_child, left\_of, right\_of,\\encloses, inside} \\
\bottomrule
\end{fitwidthtabular}
\end{table*}

\newcommand{\idsop}[1]{\begin{CJK}{UTF8}{ipxm}#1\end{CJK}}
\noindent\textbf{Graph Parsing.}
The parser converts SVG elements and their hierarchical relationships into graph nodes and edges, following the mapping in Table~\ref{tab:svg2graph}. SVG roots and groups define the graph hierarchy, while primitive elements represent geometric components. Text elements and marked primitives are retained as text and anchor nodes, respectively. Nesting, shared parentage, and anchor associations induce parent--child, sibling, and anchor-reference edges. When inferring edges from the SVG hierarchy, we connect every pair of direct child nodes of each parent, including the SVG root, with one sibling edge in each direction. A parent with $k$ direct child nodes thus contributes $k(k-1)$ directed sibling edges.

For geometric attributes, the parser composes each element's local transformation with those of its ancestors before mapping coordinates to the root \texttt{viewBox} $(v_x,v_y,v_w,v_h)$:
\begin{equation}
\widetilde x=(x-v_x)/v_w,\qquad
\widetilde y=(y-v_y)/v_h.
\end{equation}
It then extracts normalized bounding boxes, centers, widths, heights, and areas, together with up to 32 transformed sample points per primitive. This normalization preserves relative geometry independently of SVG canvas dimensions.

Semantic SVGs additionally encode domain-specific components and relations through node and edge declarations. Node attributes are mapped to numerical IDs, and typed edges specify source, target, and relation type. Unlike structural relations inferred from SVG hierarchy, semantic relations are explicitly declared and validated against the shared schema, which enforces identifier uniqueness, valid endpoints, and attribute consistency. Inverse relations must also be explicitly specified.

\begin{figure}[!htb]
    \centering
    \includegraphics[width=1.0\linewidth]{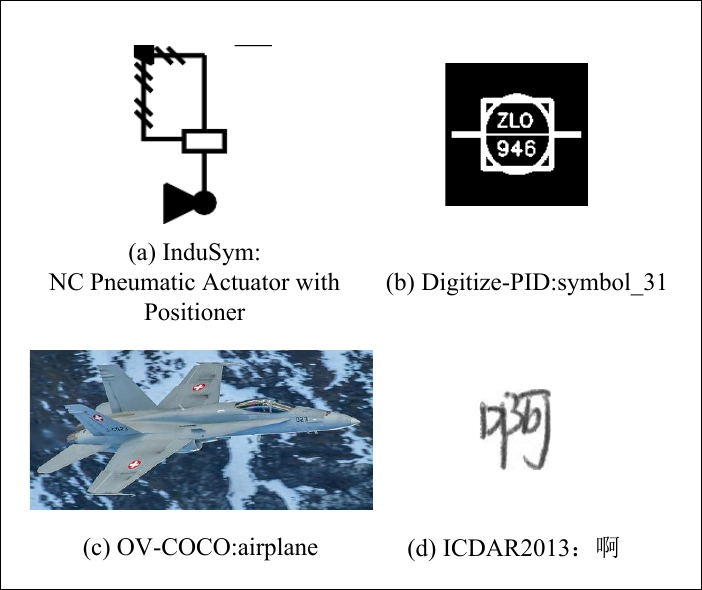}
    \caption{\textbf{Category examples} (a)--(d) in Table~\ref{tab:csg-instances} order.}
    \label{fig:semi_structured_examples}
\end{figure}

Box~\label{firstref:box:semantic-svg-aa}\ref{box:semantic-svg-aa} shows the SVG declarations for two spatial relations in the character \cjkchar{啊} (Table~\ref{tab:csg-instances} and Fig.~\ref{fig:semi_structured_examples}(d)). Its ideographic description sequence (IDS) is \mbox{\idsop{⿰}\,\cjkchar{口}\,\idsop{⿰}\,\cjkchar{阝}\,\idsop{⿹}\,\cjkchar{丁}\,\cjkchar{口}}. The excerpt shows the left component \cjkchar{口} (\textit{n002}) and the IDS operator \idsop{⿰} at the root of the right subtree (\textit{n003}). Other node attributes and rendering geometry are omitted.

The parser resolves \textit{data-src} and \textit{data-dst} against the declared node IDs, validates the relation name and ID against the domain schema, and checks that the inverse relation is declared. The two declarations produce the directed CSG edges \textit{(2, 3, left\_of)} and \textit{(3, 2, right\_of)}, where 2 and 3 are zero-based node indices in the complete graph. The airplane example in Table~\ref{tab:csg-instances} uses the same declaration format with \textit{data-src=``airplane"}, \textit{data-dst=``fuselage"}, and \textit{data-relation=``has\_part"}.

\begin{table}[!htb]
\centering
\caption{\textbf{Mapping rules from SVG elements to CSG nodes and edges.} The supported SVG primitive types are path, rect, circle, ellipse, line, polygon, and polyline.}\label{tab:svg2graph}
\footnotesize
\setlength{\tabcolsep}{3pt}
\renewcommand{\arraystretch}{1.10}
\begin{tabularx}{\linewidth}{@{}>{\raggedright\arraybackslash}p{0.25\linewidth} >{\raggedright\arraybackslash}p{0.24\linewidth} >{\raggedright\arraybackslash}X@{}}
\toprule
\multicolumn{1}{@{}>{\columncolor{headrow}[0pt][\tabcolsep]\raggedright\arraybackslash}p{0.25\linewidth}}{\tableheading{SVG construct}}
& \cellcolor{headrow}Graph element
& \multicolumn{1}{>{\columncolor{headrow}[\tabcolsep][0pt]\raggedright\arraybackslash}X@{}}{Stored attributes} \\
\midrule
svg, g & Root/group & ID, class, parent, child count \\
Primitive tag & Component & Type, normalized geometry, point count \\
text, tspan & Text node & Position, text length \\
Marked primitive & Extra anchor & Kind, position, primitive ID \\
Nesting & parent\_child & Source and target \\
Shared parent & sibling & Source and target \\
Anchor--primitive & anchor\_ref & Source and target \\
Semantic node & semantic node & Node type and domain attributes \\
Readable semantic fields & Node attributes & Role, shape and token IDs via root vocabulary \\
Typed SVG relation & Declared edge & Source, target and relation type \\
\bottomrule
\end{tabularx}
\end{table}

\begin{figure}[!htb]
\centering
\begingroup
\setlength{\fboxsep}{5pt}
\setlength{\fboxrule}{0.5pt}
\fbox{%
\begin{minipage}{\dimexpr\linewidth-2\fboxsep-2\fboxrule\relax}
\fontencoding{T1}\selectfont
\footnotesize
\setlength{\parindent}{0pt}
\begingroup
\ttfamily\raggedright\frenchspacing
\textless g id="n002" data-node-type="radical"\\
\hspace*{1em}data-token="\cjkchar{口}"/\textgreater\\
\textless g id="n003" data-node-type="operator"\\
\hspace*{1em}data-token="\idsop{⿰}"/\textgreater\par
\smallskip
\textless line class="semantic-relation"\\
\hspace*{1em}data-src="n002" data-dst="n003"\\
\hspace*{1em}data-relation="left\_of" data-relation-id="1"\\
\hspace*{1em}opacity="0"/\textgreater\\
\textless line class="semantic-relation"\\
\hspace*{1em}data-src="n003" data-dst="n002"\\
\hspace*{1em}data-relation="right\_of" data-relation-id="2"\\
\hspace*{1em}opacity="0"/\textgreater\par
\endgroup
\end{minipage}%
}
\par\smallskip
\refstepcounter{indusympromptbox}\label{box:semantic-svg-aa}%
\parbox{\linewidth}{\small\centering\textbf{Box~\theindusympromptbox.} Parsing semantic SVG relations for \cjkchar{啊}.\par}
\endgroup
\end{figure}

\noindent\textbf{Attribute Encoding.}
The parsed nodes and edges instantiate $\mathbf X_c$ and $\mathbf R_c$ in the main paper. Each node attribute $\mathbf x_i\in\mathbf X_c$ stores its node type, source identifier, class tokens, and, where applicable, normalized geometry and semantic attributes. Each edge is represented by its relation type, which determines the relation embedding $\mathbf e_{ji}$ used by SGE.

Structural relations (\textit{parent\_child}, \textit{sibling}, and \textit{anchor\_ref}) are derived from SVG hierarchy or anchor associations, whereas semantic relations follow their declared directions. Spatial relations use explicitly specified inverse edges when applicable, such as \textit{left\_of}/\textit{right\_of}, \textit{above}/\textit{below}, and \textit{encloses}/\textit{inside}. Symmetric relations such as \textit{overlaps} are encoded bidirectionally.

\section{Category Structural Graph Instances}

Table~\label{firstref:tab:csg-instances}\ref{tab:csg-instances} gives rooted category graphs from four domains, with component attributes and typed relations; Figure~\label{firstref:fig:semi_structured_examples}\ref{fig:semi_structured_examples} shows the corresponding examples.

InduSym uses nesting, same-group relations, and connection anchors, with spatial arrangement in node geometry. Digitize-PID symbol\_31 records two connection anchors. Natural-object definitions supply part and attribute nodes; airplane has five parts and two attributes. Chinese ideographic descriptions supply glyph parts and spatial operators, with relations such as left\_of and encloses.

\begin{figure}[!htb]
    \centering
    \includegraphics[width=\linewidth]{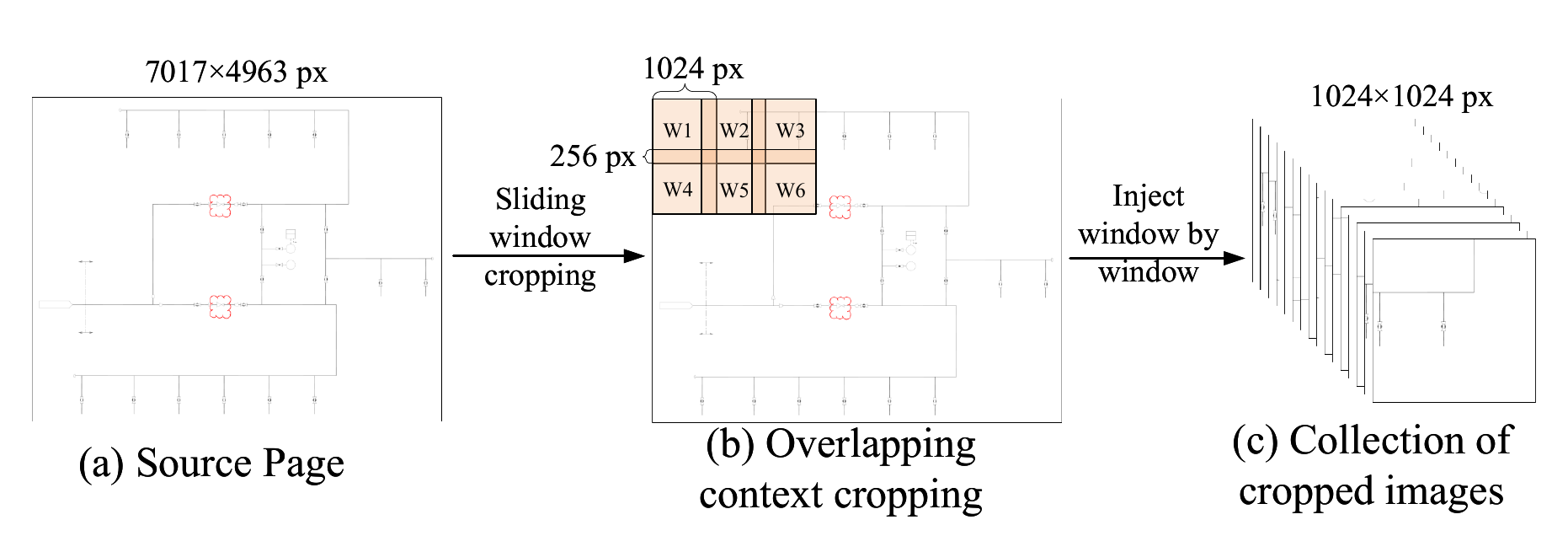}
    \caption{\textbf{Overlapping-window cropping for InduSym.}}\label{fig:indusym-pipeline}
  \end{figure}

\section{Benchmark Statistics}

\begin{table}[!htb]
\centering
\caption{\textbf{InduSym annotations and 5-shot training supervision.}}\label{tab:indusym-benchmark}
\footnotesize
\setlength{\tabcolsep}{2.5pt}
\begin{tabularx}{\linewidth}{X c c c c}
\toprule
\headband
\begin{tabular}[c]{@{}l@{}}\tableheading{Category group}\end{tabular} &
\begin{tabular}[c]{@{}c@{}}Classes\end{tabular} &
\begin{tabular}[c]{@{}c@{}}Instances\\ (train / val / test)\end{tabular} &
\begin{tabular}[c]{@{}c@{}}Source\\ (total)\end{tabular} &
\begin{tabular}[c]{@{}c@{}}Positive\\ supervision\end{tabular} \\
\midrule
Base & 204 & 71,657 / 7,934 / 8,261 & 87,852 & 71,657 \\
Few-shot & 45 & 552 / 142 / 168 & 862 & 225 \\
Zero-shot & 26 & 0 / 112 / 86 & 198 & 0 \\
\midrule
\ours \textbf{Total} & 275 & 72,209 / 8,188 / 8,515 & 88,912 & 71,882 \\
\bottomrule
\end{tabularx}
\end{table}

Table~\label{firstref:tab:indusym-benchmark}\ref{tab:indusym-benchmark} separates the available annotations from the positive supervision used for training. The dataset contains 88,912 instances, but only five support instances per few-shot category are used in the 5-shot setting, giving 225 few-shot positives. Zero-shot categories supply no positive training instances. Figure~\label{firstref:fig:indusym-pipeline}\ref{fig:indusym-pipeline} illustrates overlapping-window cropping.

\section{Additional Experiments}

\subsection{Structural and Verification Ablation Settings}

In Table~\ref{tab:composition-ablation}, checkmarks denote retained modules, not guaranteed nonempty retrieval or executed refinement rounds.

\noindent\textbf{Empty input.} No category-specific CSG nodes or edges are supplied. In Phase II, each category retains its text prompt and receives the same one-node placeholder graph with zero-valued attributes and no edges. The cached structure embedding is set to zero. SGE and MSQ treat the placeholder as a valid node, text fusion and the adapter remain active, and refinement follows the usual gating rules.

w/o relations retains correct nodes and attributes without edges; Wrong relations permutes edge endpoints. Edge changes affect SGE message passing, MSQ smoothing, and NRM redundancy masks, so these comparisons assess relation use across the system.

Both Figure~\ref{fig:verification} variants retain two refinement rounds.

\noindent\textbf{w/o NRM: structure-only correction.} The candidate's predicted class ID selects structural queries and a valid-node mask. The expert receives their masked mean $\bar{\mathbf h}_{c_k}=|\mathcal V_{c_k}^{+}|^{-1}\sum_{i\in\mathcal V_{c_k}^{+}}\mathbf h_{c_k,i}^{+}$ with zero visual input. Local visual extraction and the matching/selection/filtering steps are bypassed. Reliability is the predicted-class detector score. Correction and downstream feedback remain active.

\noindent\textbf{w/o LSF: global broadcast feedback.} NRM and local scatter remain active. Each scale's residual is spatially averaged over all $H_\ell W_\ell$ positions and broadcast globally; reliability/ambiguity maps become $\mathbf S^{\rho}=\mathbf 1$ and $\mathbf S^{m}=\mathbf 0$. Subsequent SCM gates, projections, and clipping remain active. This jointly changes candidate support, map conditioning, and residual magnitude; it does not isolate locality or remove the learned gates.

\subsection{Update Strategies}

\begin{table}[!htb]
\centering
\caption{\textbf{Structure and detector updates.} $\checkmark$: present/trainable.}\label{tab:update-strategies}
\footnotesize
\setlength{\tabcolsep}{2.3pt}
\begin{tabularx}{\linewidth}{X ccc ccc}
\toprule
\headband \makecell[cl]{\tableheading{Method}} & \makecell[c]{Struct.\\branch} & \makecell[c]{Struct.\\update} & \makecell[c]{Det.\\update} & \makecell[c]{\APb} & \makecell[c]{\APf} & \Hbf \\
\midrule
Baseline & \na & \na & \checkmark & 19.07 & \second{31.70} & \second{23.81} \\
w/o Detector update & \checkmark & \checkmark & \na & \second{22.38} & 15.06 & 18.00 \\
w/o Structure update & \checkmark & \na & \checkmark & 18.88 & 25.59 & 21.72 \\
\ours \textbf{Struct2Det (ours)} & \checkmark & \checkmark & \checkmark & \best{36.60} & \best{37.72} & \best{37.15} \\
\bottomrule
\end{tabularx}
\end{table}

Table~\label{firstref:tab:update-strategies}\ref{tab:update-strategies} compares updates, with freezing in both phases. Structure-only updates raise \APb{} but lower \APf{} relative to YOLO-World; detector-only updates lower both. Joint updates give the best balance. Individual modules are not isolated.

\subsection{Support-Set Size}
\begin{table}[!htb]
\centering
\caption{\textbf{Effect of support-set size} on InduSym and Digitize-PID, using 1, 5, and 10 instances per few-shot category.}\label{tab:support-scale}
\footnotesize
\setlength{\tabcolsep}{5pt}
\begin{tabularx}{\linewidth}{l *{3}{Y} c *{3}{Y}}
\toprule
 & \multicolumn{3}{c}{\tableheading{InduSym}} & & \multicolumn{3}{c}{\tableheading{Digitize-PID}} \\
\cmidrule(lr){2-4}\cmidrule(l){6-8}
\headband \tableheading{Shots} & \APb & \APf & \Hbf & & \APb & \APf & \Hbf \\
\midrule
1 & 35.85 & 25.60 & 29.87 & & 92.91 & 33.37 & 49.10 \\
5 & 36.60 & 37.72 & 37.15 & & 94.37 & 64.48 & 76.61 \\
10 & 35.75 & 39.75 & 37.64 & & 95.48 & 73.68 & 83.18 \\
\bottomrule
\end{tabularx}
\end{table}

Table~\label{firstref:tab:support-scale}\ref{tab:support-scale} shows increasing \APf{} and \Hbf{} with support. Gains in \APf{} shrink from 12.12 to 2.03 points on InduSym and 31.11 to 9.20 on Digitize-PID across the two shot increments. At 10 shots, InduSym loses 0.85 points in \APb{}, while Digitize-PID improves both groups.

\subsection{Computational Cost}
\noindent\textbf{Measurement protocol.}
We measured computational cost on an NVIDIA A800 GPU using
\(640\times640\) detector inputs, a batch size of 1, and FP32 precision.
For Struct2Det, we used a 275-class vocabulary. With the standard inference gate enabled, every evaluated image completed the maximum of two refinement rounds. SGE outputs and image-independent structural queries were cached and reused across images.

\begin{table}[!htb]
\centering
\caption{\textbf{Computational cost.} Parameters, FLOPs, and per-image inference time at $640\times640$ resolution.}\label{tab:complexity}
\footnotesize
\setlength{\tabcolsep}{4pt}
\renewcommand{\arraystretch}{1.16}
\begin{tabularx}{\linewidth}{X rrr}
\toprule
\headband \tableheading{Method} & Params (M) & FLOPs (G) & Time (s) \\
\midrule
RegionCLIP & 120.268 & 2462.319 & 0.197 \\
CORA & 190.296 & 35547.379 & 0.383 \\
YOLO-World & 76.188 & 1319.376 & 0.141 \\
Grounding DINO & 172.840 & 360.749 & 1.225 \\
SSDPID & 26.790 & 10.353 & 0.015 \\
YOLOE & 11.232 & 28.966 & 0.009 \\
HD-OVD & 146.269 & 224.415 & 0.055 \\
\ours \textbf{Struct2Det (ours)} & 107.018 & 1809.646 & 0.234 \\
\bottomrule
\end{tabularx}
\end{table}

Relative to YOLO-World, Table~\label{firstref:tab:complexity}\ref{tab:complexity} shows 40.5\% more parameters, and 37.2\% more FLOPs, and 66.0\% longer inference. Graph-level SGE, fixed-count MSQ sampling, and bounded NRM matching constrain computation; the backbone and dense SCM gates remain resolution-dependent.

\subsection{Few-Shot Detection Analysis}

We analyze few-shot detection on InduSym through local component matching, discrimination between visually similar classes, and a qualitative comparison on an engineering drawing.

\begin{figure}[!t]
\centering
\begin{minipage}[t]{\dimexpr(\linewidth-\columnsep)/2\relax}
  \vspace{0pt}
  \captionsetup{type=figure}
  \centering
  \includegraphics[width=\linewidth]{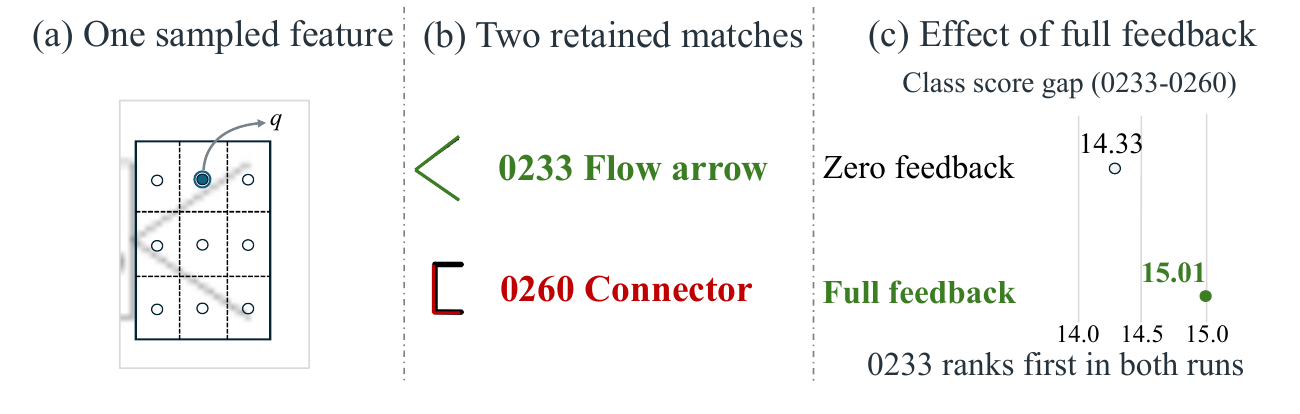}
  \caption{\textbf{Local matching and structural feedback for a few-shot flow arrow.}}
\label{fig:few-shot-local-supple}
\end{minipage}%
\hfill
\begin{minipage}[t]{\dimexpr(\linewidth-\columnsep)/2\relax}
  \vspace{0pt}
  \captionsetup{type=figure}
  \centering
  \includegraphics[width=\linewidth]{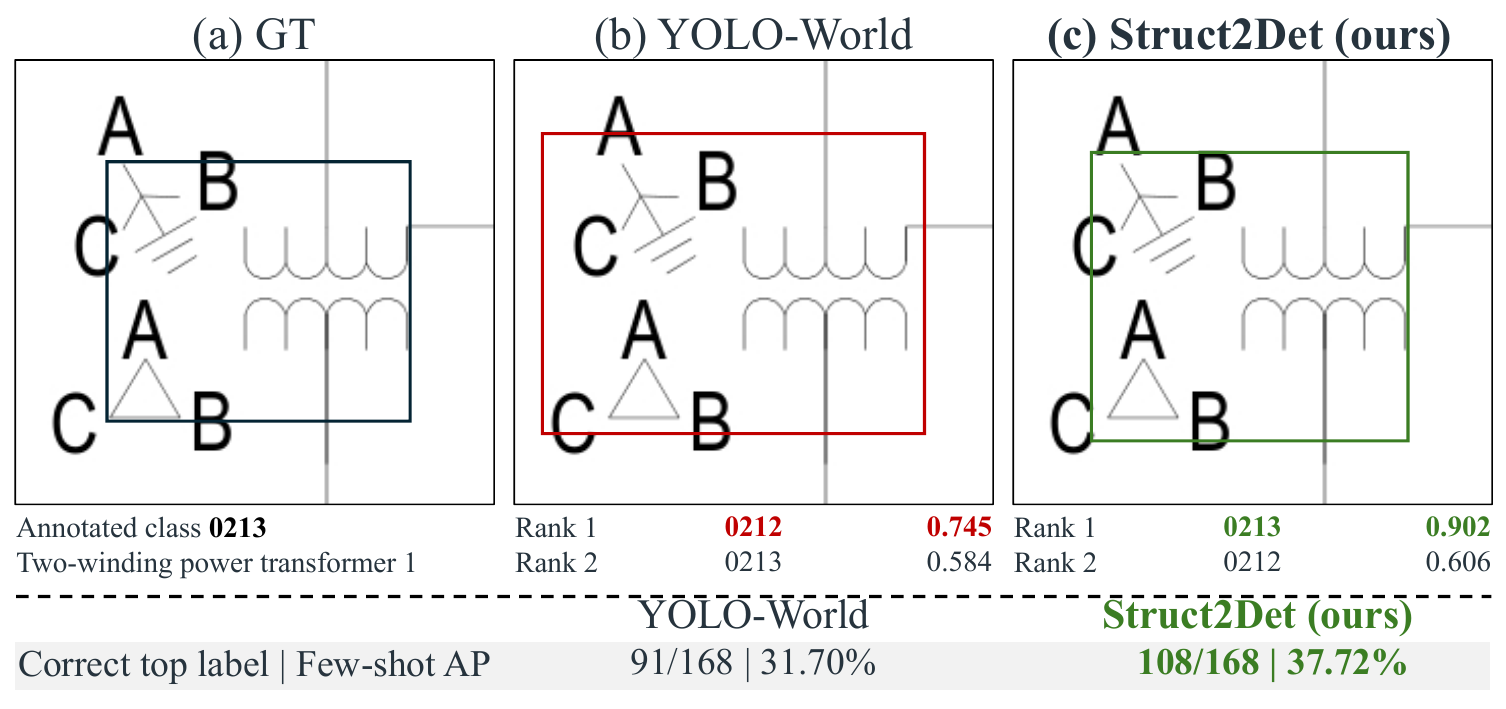}
  \caption{\textbf{Discrimination between visually similar few-shot symbols.}}
\label{fig:few-shot-discrimination-supple}
\end{minipage}
\par\medskip
\includegraphics[width=\textwidth]{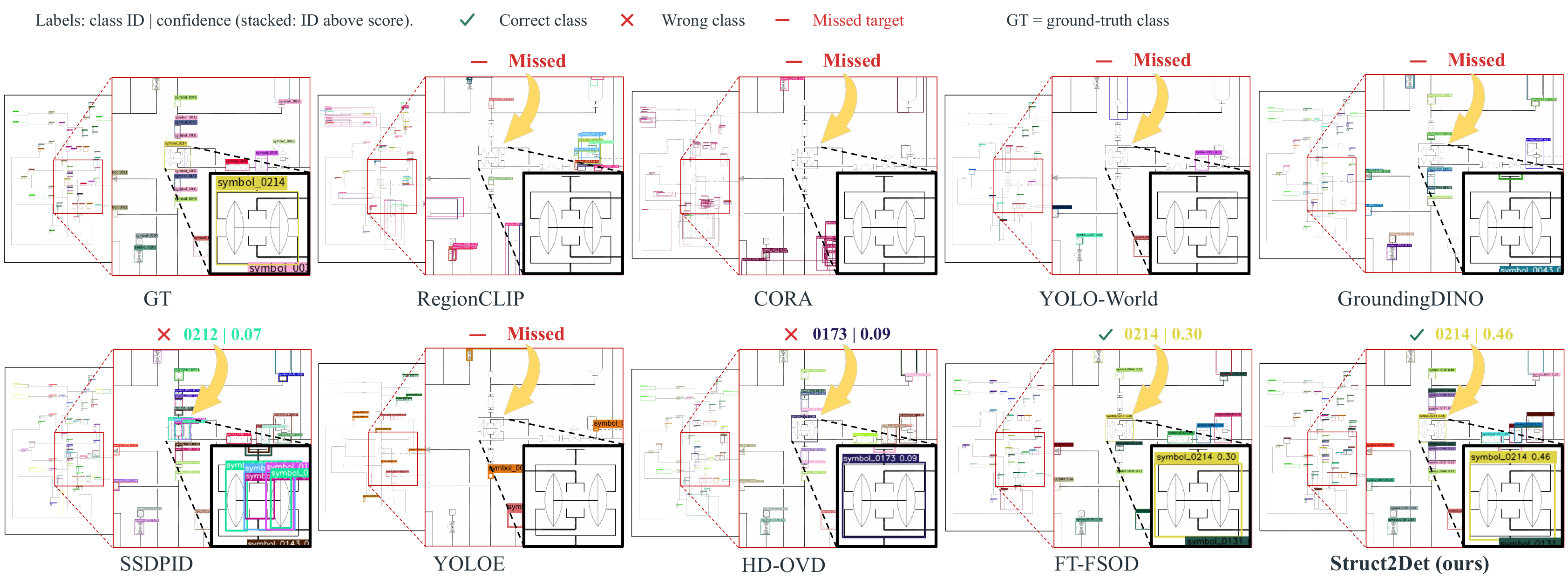}
\caption{\textbf{Qualitative example of few-shot detection Analysis.}
The insets highlight a double-diaphragm pump (\textit{symbol\_0214}), a category with five labeled support instances, on a drawing from the InduSym. Struct2Det correctly classifies and localizes the pump with a detection score of 0.46 and an IoU of 0.92. FT-FSOD also detects the target (score 0.30), while several baselines miss it or predict a different class in the highlighted region.}\label{fig:qualitative_InduSym-few-shot_supple}
\end{figure}

\noindent\textbf{Local matching and feedback.}
Figure~\ref{fig:few-shot-local-supple}(a,b) shows how one feature sampled inside a flow-arrow candidate matches a graph node from each of two class hypotheses, flow arrow (0233) and connector (0260). With the candidate and classification head fixed, full feedback widens the flow arrow's lead from 14.33 to 15.01 percentage points (c). The flow arrow ranks first in both settings.

\noindent\textbf{Discrimination between similar symbols.}
In Figure~\ref{fig:few-shot-discrimination-supple}, YOLO-World ranks the visually similar transformer class 0212 first, whereas Struct2Det ranks the annotated class 0213 first. Across 168 few-shot instances on 68 test pages, Struct2Det assigns the correct top label to 108 instances and YOLO-World to 91.

\noindent\textbf{Detection on an engineering drawing.}
Figure~\ref{fig:qualitative_InduSym-few-shot_supple} compares detectors on a double-diaphragm pump (\textit{symbol\_0214}). Struct2Det correctly detects the pump with an IoU of 0.92. FT-FSOD is the only baseline shown that detects it with the correct class.

\clearpage
\onecolumn
\raggedbottom
\ifdefined\leftlinenumbers\leftlinenumbers*\fi

\subsection{More Qualitative Results}

\begin{figure}[!p]
\centering
\includegraphics[width=\textwidth]{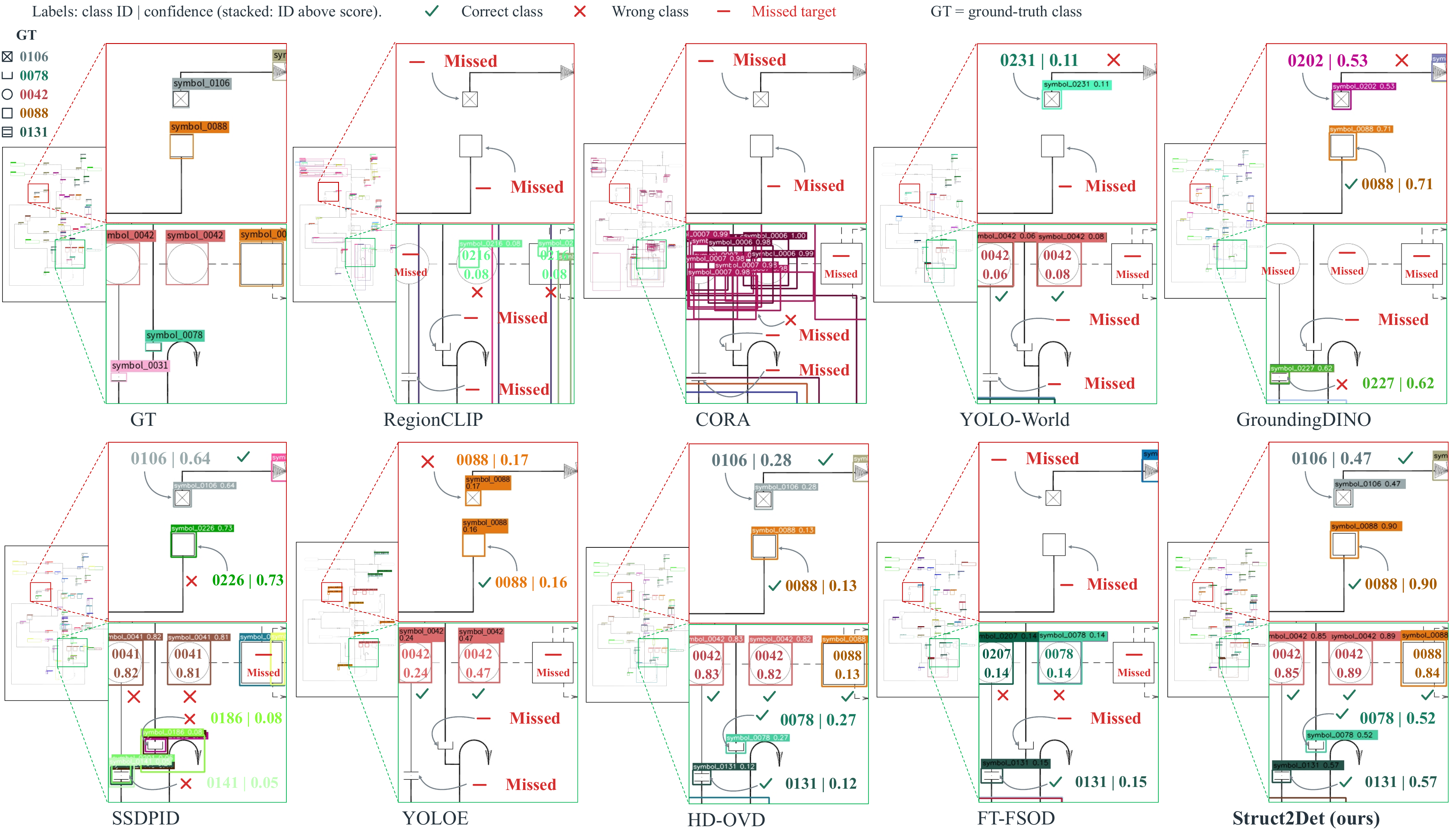}
\caption{\textbf{Full qualitative comparison on InduSym.}
This figure provides the complete version of
Fig.~\ref{fig:qualitative_InduSym}, covering all compared methods.
Red and green frames mark enlarged regions for comparing missed detections,
extra detections, and category confusion. Struct2Det detects all seven highlighted targets, achieving the highest average confidence.}\label{fig:qualitative_InduSym_supple}
\end{figure}

\begin{figure}[!p]
\centering
\includegraphics[width=\textwidth]{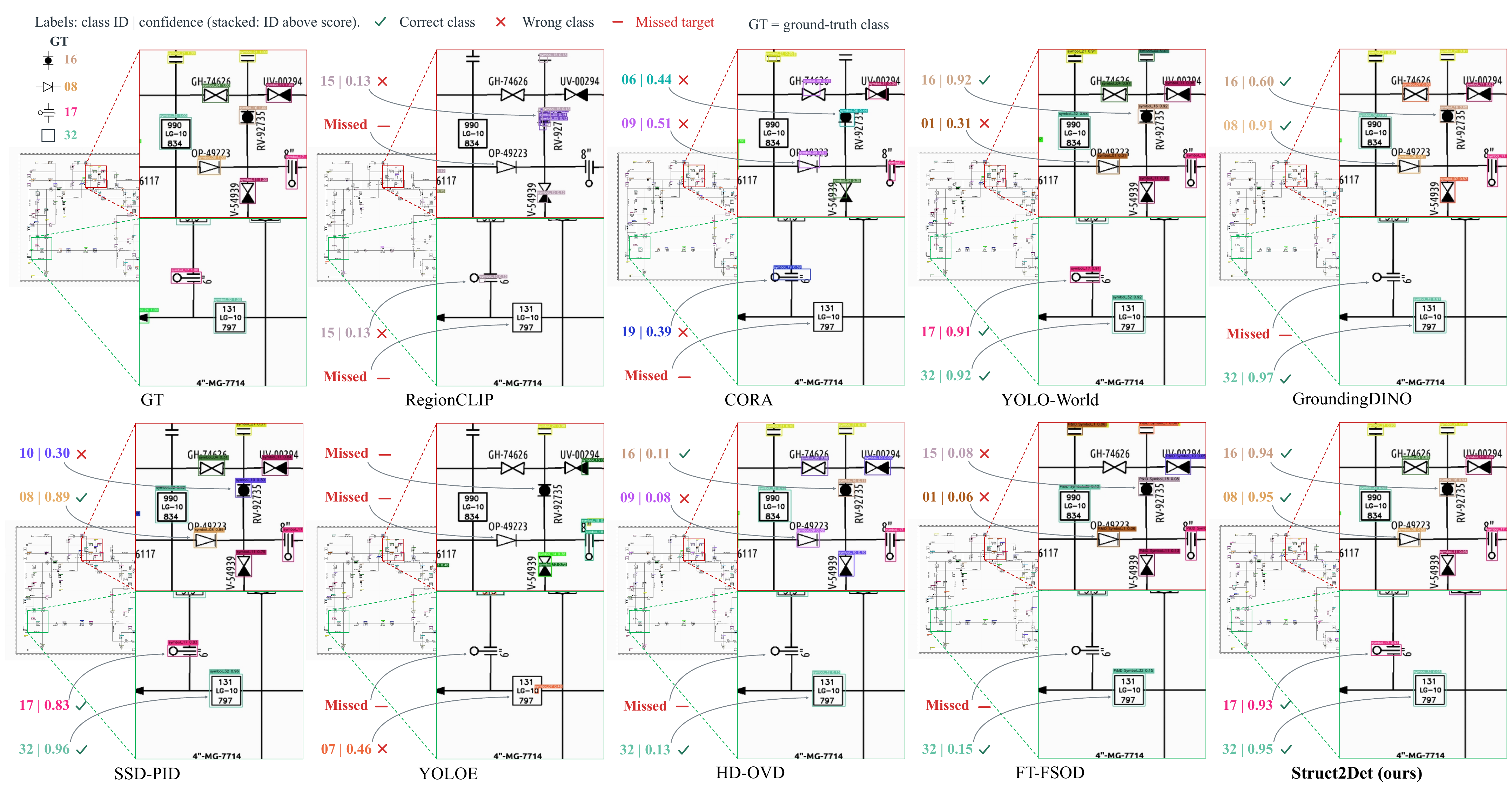}
\caption{\textbf{Qualitative comparison on Digitize-PID.}
    The red and green boxes indicate the source regions of the enlarged crops. Colored labels mark the four ground-truth symbols (IDs 16, 08, 17, and 32), while check marks, crosses, and dashes denote correct classifications, misclassifications, and missed detections, respectively. Struct2Det correctly identifies all four symbols, demonstrating its robustness to dense piping and visually similar categories.}\label{fig:qualitative_Digitize-PID}
\end{figure}

\begin{figure}[!p]
\centering
\includegraphics[width=\textwidth]{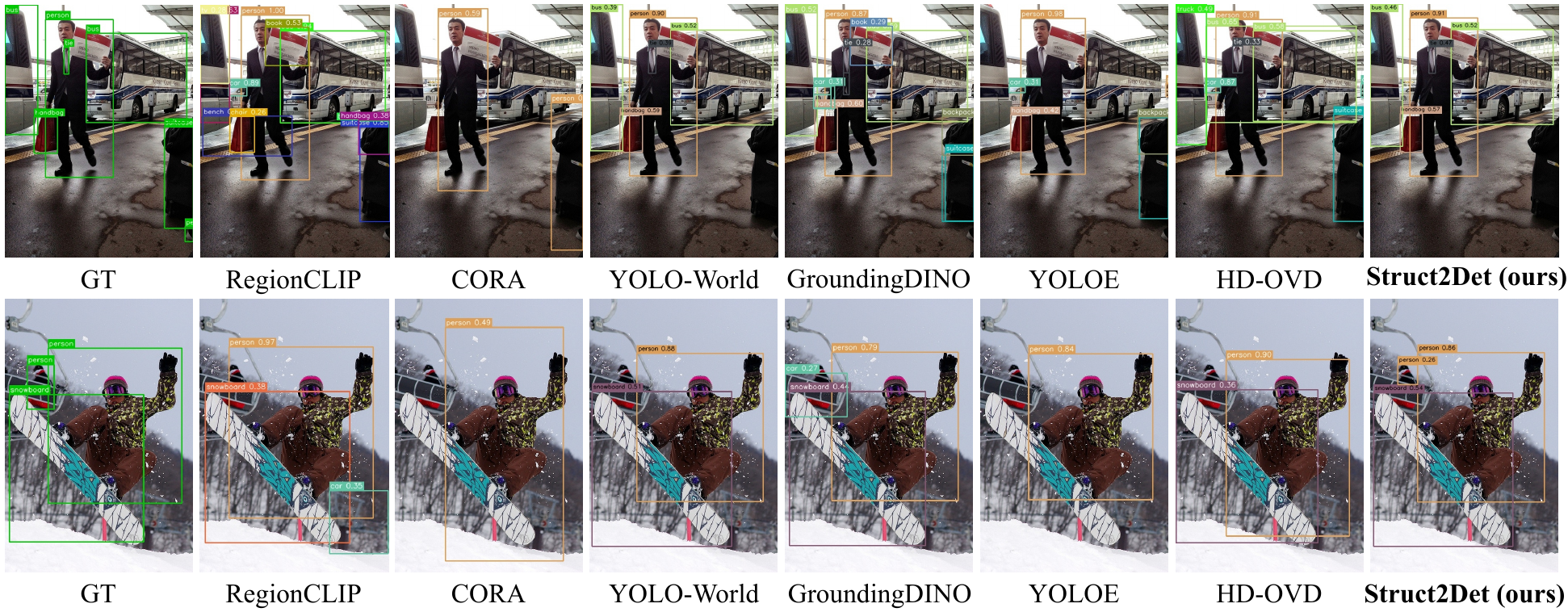}
\caption{\textbf{Qualitative comparison on OV-COCO.}
    The two rows show COCO val2017 images 202339 and 491130 under the 48-base/17-novel split. RegionCLIP, CORA, and HD-OVD use released checkpoints. The HD-OVD panel uses AdaMixer. In the top row, Struct2Det detects both buses and the central pedestrian. In the bottom row, YOLO-World detects the snowboard (0.51) but misses the person in the background; Struct2Det detects both people and the snowboard.}\label{fig:qualitative_OV-COCO}
\end{figure}

\begin{figure}[!p]
\centering
\includegraphics[width=\textwidth]{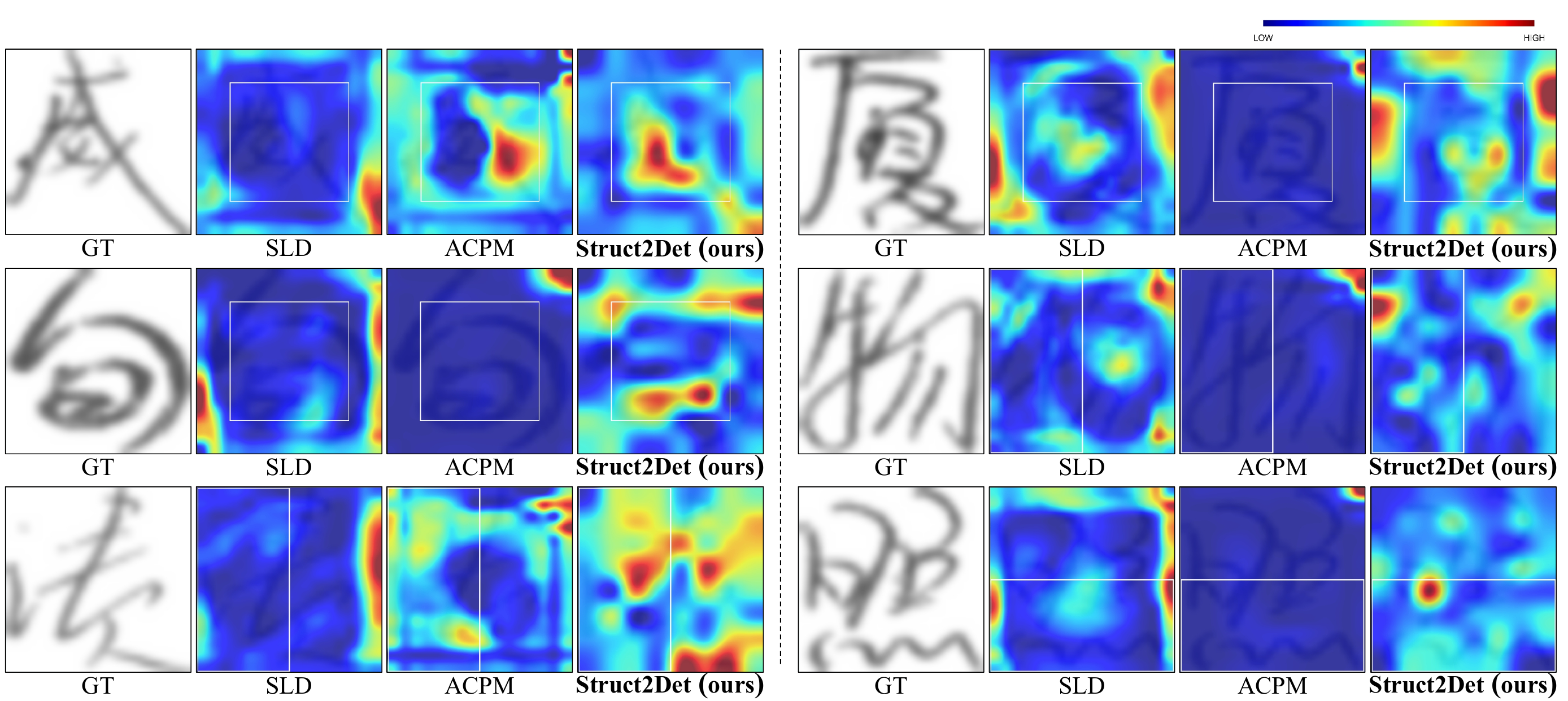}
\caption{\textbf{Structural CAMs and recognition results on ICDAR2013.}
Heatmaps show independently normalized averages of ground-truth-conditioned structural-unit CAMs. Struct2Det CAMs are extracted from the graph branch before support fusion. Values indicate the percentage of CAM mass within the white boxes, in the order SLD, ACPM, and Struct2Det.
\textbf{Left:} The highlighted components correspond to \cjkchar{威}, \cjkchar{旬}, and \cjkchar{渍}. SLD and ACPM misclassify all three, whereas Struct2Det recognizes them correctly. Their in-box CAM percentages are (17.1, 42.6, 59.8), (21.2, 17.2, 60.9), and (35.0, 57.5, 43.8), respectively.
\textbf{Right:} All methods correctly recognize \cjkchar{厦}, \cjkchar{物}, and \cjkchar{照}. Their in-box CAM percentages are (28.9, 26.9, 45.2), (41.1, 20.4, 46.4), and (48.2, 38.4, 51.2), respectively.}
\label{fig:qualitative_ICDAR}
\end{figure}

Figures~\ref{fig:qualitative_InduSym_supple}, \ref{fig:qualitative_Digitize-PID}, \ref{fig:qualitative_OV-COCO}, and \ref{fig:qualitative_ICDAR} present qualitative comparisons on InduSym, Digitize-PID, OV-COCO, and ICDAR, respectively.


\end{document}